\pdfoutput=1   
\documentclass[twocolumn]{article}
\usepackage{preprint}   

\usepackage[T1]{fontenc}
\usepackage{natbib}

\usepackage{stfloats}
\usepackage{amsmath,amsfonts,bm}

\def\eqref#1{equation~\ref{#1}}

\def\1{\bm{1}}

\DeclareMathAlphabet{\mathsfit}{\encodingdefault}{\sfdefault}{m}{sl}
\SetMathAlphabet{\mathsfit}{bold}{\encodingdefault}{\sfdefault}{bx}{n}

\def\eqref#1{(\ref{#1})}

\usepackage{graphicx}
\usepackage{booktabs}
\usepackage{amsmath,amssymb}
\usepackage{amsthm}
\newtheorem{proposition}{Proposition}
\theoremstyle{definition}
\newtheorem{assumption}{Assumption}
\newtheorem{definition}{Definition}
\theoremstyle{plain}
\usepackage[most]{tcolorbox}
\definecolor{boxorange}{HTML}{C98526}\definecolor{boxorangebg}{HTML}{FBF4E9}
\definecolor{boxgrey}{HTML}{8E8E8E}\definecolor{boxgreybg}{HTML}{F3F3F3}
\definecolor{boxblue}{HTML}{2F6DAE}\definecolor{boxbluebg}{HTML}{EBF2FA}
\tcbset{revarbox/.style={enhanced, breakable, frame hidden, boxrule=0pt, sharp corners, boxsep=0pt,
  left=6pt, right=4pt, top=3pt, bottom=3pt}}
\tcolorboxenvironment{assumption}{revarbox, colback=boxorangebg, borderline west={2pt}{0pt}{boxorange}}
\tcolorboxenvironment{proposition}{revarbox, colback=boxgreybg, borderline west={2pt}{0pt}{boxgrey}}
\tcolorboxenvironment{definition}{revarbox, colback=boxbluebg, borderline west={2pt}{0pt}{boxblue}}
\usepackage{multirow}
\usepackage{array}
\usepackage{subcaption}
\usepackage{placeins}
\usepackage[ruled,vlined,linesnumbered,norelsize]{algorithm2e}
\usepackage{xcolor}
\usepackage[breaklinks=true]{hyperref}
\usepackage{url}

\graphicspath{{figures/}}

\newcommand{\methodname}{ReVar\-3R}
\newcommand{\down}{$\downarrow$}
\newcommand{\up}{$\uparrow$}
\newcommand{\Rthree}{\ensuremath{\mathrm{Sim}(3)}}

\newcommand{\pithree}{\ensuremath{\pi^3}}

\title{\methodname{}: Gauge-Aware Perturbation Uncertainty for Feed-Forward 3D Reconstruction}

\author{
  Sammam Mahdi$^{1,*}$, \ Fariha Binta Salim$^{1}$, \ Rakin Bin Rabbani$^{1}$, \
  Aniqua Nusrat Zereen$^{2}$ \\[2pt]
  \normalfont\normalsize $^{1}$BRAC University \qquad $^{2}$Mahidol University \qquad
  $^{*}$Corresponding author \\[2pt]
  \normalfont\small\texttt{\{sammam.mahdi, fariha.binta.salim\}@g.bracu.ac.bd} \quad
  \normalfont\small\texttt{rakin.rabbani@bracu.ac.bd} \\[1pt]
  \normalfont\small\texttt{aniqua.zer@mahidol.ac.th}
}
\date{}   

\newcommand{\tblaggabl}{%
Trace variance (used) & \textbf{0.359} & \textbf{0.464} & \textbf{0.394} \\
Median absolute deviation & 0.440 & 0.507 & 0.462 \\
Largest eigenvalue & 0.365 & 0.468 & 0.399 \\%
}

\newcommand{\tblalignabl}{%
group & none & 0.543 & 0.526 & 0.578 & $+0.20$ & $3/18$ \\
 & translation & 0.500 & 0.459 & 0.581 & $+0.27$ & $8/18$ \\
 & rigid $\mathrm{SE}(3)$ & 0.460 & 0.420 & 0.539 & $+0.31$ & $7/18$ \\
 & similarity $\mathrm{Sim}(3)$ & \textbf{0.392} & \textbf{0.347} & 0.483 & $\mathbf{+0.36}$ & $\mathbf{11/18}$ \\
\midrule
reference & run $0$ & 0.392 & 0.347 & 0.483 & $+0.36$ & $11/18$ \\
 & medoid & 0.395 & 0.351 & 0.483 & $+0.37$ & $11/18$ \\
 & random & 0.394 & 0.349 & 0.484 & $+0.36$ & $11/18$ \\
\midrule
scope & per-view & \textbf{0.392} & \textbf{0.347} & 0.483 & $\mathbf{+0.36}$ & $\mathbf{11/18}$ \\
 & scene-level & 0.446 & 0.404 & 0.530 & $+0.26$ & $7/18$ \\%
}

\newcommand{\tblaurc}{%
7-Scenes & 0.249 & 0.189 & 0.224 & \textbf{0.179} & 0.127 & 0.068 & 0.102 & \textbf{0.067} \\
TUM      & 0.079 & 0.082 & 0.101 & \textbf{0.067} & 0.027 & 0.030 & 0.050 & \textbf{0.019} \\
ETH3D    & 3.812 & 3.535 & \textbf{3.408} & 3.513 & 2.114 & 1.837 & \textbf{1.710} & 1.834 \\
DTU      & 34.95 & \textbf{25.93} & 33.20 & 33.59 & 16.83 & \textbf{7.81} & 15.08 & 17.87 \\
KITTI    & 2.929 & \textbf{2.418} & 3.524 & 2.833 & 1.231 & \textbf{0.721} & 1.827 & 1.198 \\%
}

\newcommand{\tblseedgrid}{%
MASt3R & BlendedMVS & 0.282 & $0.349\,{\pm}\,0.011$ & 0/5 \\
MASt3R & DTU & 0.463 & $0.330\,{\pm}\,0.012$ & 5/5 \\
MASt3R & ETH3D & 0.570 & $0.556\,{\pm}\,0.005$ & 5/5 \\
MASt3R & KITTI & 0.358 & $0.240\,{\pm}\,0.007$ & 5/5 \\
MASt3R & 7-Scenes & 0.587 & $0.353\,{\pm}\,0.005$ & 5/5 \\
MASt3R & TUM & 0.231 & $0.172\,{\pm}\,0.004$ & 5/5 \\
\pithree{} & BlendedMVS & 0.187 & $0.296\,{\pm}\,0.004$ & 0/5 \\
\pithree{} & DTU & 0.273 & $0.230\,{\pm}\,0.007$ & 5/5 \\
\pithree{} & ETH3D & 0.528 & $0.593\,{\pm}\,0.003$ & 0/5 \\
\pithree{} & KITTI & 0.387 & $0.265\,{\pm}\,0.011$ & 5/5 \\
\pithree{} & 7-Scenes & 0.517 & $0.363\,{\pm}\,0.005$ & 5/5 \\
\pithree{} & TUM & 0.246 & $0.181\,{\pm}\,0.002$ & 5/5 \\
VGGT & BlendedMVS & 0.786 & $0.387\,{\pm}\,0.014$ & 5/5 \\
VGGT & DTU & 0.248 & $0.172\,{\pm}\,0.002$ & 5/5 \\
VGGT & ETH3D & 0.740 & $0.628\,{\pm}\,0.005$ & 5/5 \\
VGGT & KITTI & 0.479 & $0.242\,{\pm}\,0.005$ & 5/5 \\
VGGT & 7-Scenes & 0.712 & $0.486\,{\pm}\,0.006$ & 5/5 \\
VGGT & TUM & 0.308 & $0.191\,{\pm}\,0.005$ & 5/5 \\%
}

\newcommand{\tblalignedmc}{%
MASt3R & DTU & 0.463 & 0.491 & 0.358 & \textbf{0.343} \\
MASt3R & ETH3D & 0.570 & 0.568 & \textbf{0.540} & 0.576 \\
MASt3R & 7-Scenes & 0.587 & 0.494 & \textbf{0.374} & 0.389 \\
\pithree{} & DTU & 0.273 & 0.613 & 0.272 & \textbf{0.269} \\
\pithree{} & ETH3D & 0.528 & 0.490 & \textbf{0.453} & 0.481 \\
\pithree{} & 7-Scenes & 0.517 & 0.547 & 0.412 & \textbf{0.409} \\
VGGT & DTU & 0.248 & 0.315 & \textbf{0.230} & 0.236 \\
VGGT & ETH3D & 0.740 & \textbf{0.507} & 0.576 & 0.560 \\
VGGT & 7-Scenes & 0.712 & 0.610 & \textbf{0.522} & 0.526 \\%
}

\newcommand{\tblcoredecomp}{%
VGGT & 7-Scenes & 0.712 & 0.655 & \textbf{0.536} & 0.491 \\
VGGT & DTU & 0.248 & 0.313 & \textbf{0.236} & 0.171 \\
VGGT & ETH3D & 0.740 & 0.520 & \textbf{0.578} & 0.635 \\
VGGT & BlendedMVS & 0.786 & 0.597 & \textbf{0.276} & 0.407 \\
\pithree{} & 7-Scenes & 0.517 & 0.541 & \textbf{0.439} & 0.363 \\
\pithree{} & DTU & 0.273 & 0.660 & 0.276 & 0.220 \\
\pithree{} & ETH3D & 0.528 & 0.487 & \textbf{0.522} & 0.593 \\
\pithree{} & BlendedMVS & 0.187 & 0.243 & 0.251 & 0.298 \\
MASt3R & 7-Scenes & 0.587 & 0.479 & \textbf{0.391} & 0.358 \\
MASt3R & DTU & 0.463 & 0.500 & \textbf{0.364} & 0.350 \\
MASt3R & ETH3D & 0.570 & 0.645 & 0.595 & 0.558 \\
MASt3R & BlendedMVS & 0.282 & 0.843 & 0.659 & 0.367 \\%
}

\newcommand{\tblnewds}{%
TUM & VGGT & 0.296 / +0.48 & 0.819 / $-$0.01 & 0.446 / +0.39 & \textbf{0.199} / +0.51 \\
TUM & \pithree{} & 0.273 / +0.25 & 0.286 / +0.65 & \textbf{0.212} / +0.51 & \textbf{0.198} / +0.46 \\
TUM & MASt3R & 0.226 / +0.63 & 0.433 / +0.24 & 0.270 / +0.44 & \textbf{0.182} / +0.56 \\
\midrule
KITTI & VGGT & 0.401 / +0.30 & 0.592 / +0.32 & 0.432 / +0.58 & \textbf{0.237} / +0.62 \\
KITTI & \pithree{} & 0.385 / +0.45 & 0.451 / +0.53 & \textbf{0.337} / +0.64 & \textbf{0.273} / +0.74 \\
KITTI & MASt3R & 0.346 / +0.53 & 0.477 / +0.45 & \textbf{0.313} / +0.67 & \textbf{0.215} / +0.77 \\%
}

\newcommand{\tblperscene}{%
VGGT & 7-Scenes & 0.712 & 0.491 & 0.06 & 0.32 \\
VGGT & DTU & 0.248 & 0.171 & 0.31 & 0.44 \\
VGGT & ETH3D & 0.740 & 0.635 & $-0.11$ & 0.13 \\
VGGT & BlendedMVS & 0.786 & 0.407 & $-0.16$ & 0.43 \\
\pithree{} & 7-Scenes & 0.517 & 0.363 & 0.36 & 0.48 \\
\pithree{} & DTU & 0.273 & 0.220 & 0.37 & 0.47 \\
\pithree{} & ETH3D & 0.528 & 0.593 & 0.19 & 0.19 \\
\pithree{} & BlendedMVS & 0.187 & 0.298 & 0.59 & 0.48 \\
MASt3R & 7-Scenes & 0.587 & 0.358 & 0.04 & 0.48 \\
MASt3R & DTU & 0.463 & 0.350 & 0.13 & 0.24 \\
MASt3R & ETH3D & 0.570 & 0.558 & 0.27 & 0.22 \\
MASt3R & BlendedMVS & 0.282 & 0.367 & 0.50 & 0.35 \\%
}

\newcommand{\tblposes}{%
\pithree{} & 7-Scenes & 0.217 & \textbf{0.175} & 0.276 & 0.351 & 0.299 & 0.314 & $0.38^{\ast}$ \\
\pithree{} & DTU & 0.210 & \textbf{0.119} & $-0.006$ & 0.521 & 0.156 & \textbf{0.088} & $0.15$ \\
\pithree{} & ETH3D & 0.534 & \textbf{0.519} & 0.157 & 0.007 & 0.401 & 0.445 & $0.07$ \\
VGGT & 7-Scenes & 0.556 & \textbf{0.461} & $-0.297$ & $-0.301$ & 0.322 & \textbf{0.291} & $0.70^{\ast}$ \\
VGGT & DTU & 0.083 & \textbf{0.048} & 0.121 & 0.386 & 0.355 & \textbf{0.305} & $0.66^{\ast}$ \\
VGGT & ETH3D & 0.714 & \textbf{0.569} & $-0.195$ & $-0.233$ & 0.655 & \textbf{0.507} & $0.61^{\ast}$ \\%
}

\newcommand{\tblprobe}{%
Built-in conf. & 0.702 / 0.09 & 0.402 / 0.24 & 0.754 / $-0.09$ \\
Learned probe (frozen features) & 0.791 / 0.09 & 0.245 / 0.56 & 0.644 / 0.14 \\
\methodname{} (aligned perturbation) & \textbf{0.546} / 0.29 & \textbf{0.170} / 0.46 & \textbf{0.510} / 0.15 \\%
}

\begin{document}

\twocolumn[
  \begin{@twocolumnfalse}
\maketitle
\begin{abstract}
A correctly reconstructed distant point appears uncertain even when a frozen 3D model processes equivalent
inputs because its output frame rotates fractionally. This exposes a weakness of training-free perturbation
uncertainty: when outputs contain an unobserved symmetry, run-to-run variation potentially reflects symmetry rather than
error. Existing alternatives have trade-offs:
built-in confidence is outperformed in most evaluated conditions, while trained evidential heads require
model-specific supervision. For point maps, this research derives a closed-form, error-independent variance term
that grows with scene extent and potentially overwhelms the desired signal. Simulation reproduces the effect; all $30$ real
VGGT view-sets tested exhibit its predicted $\|x_p\|^2$ signature. \methodname{} robustly registers predictions to
a common similarity frame before computing per-point variance, without retraining or modifying the frozen model. Optional calibration and
fusion use a held-out split. Across VGGT, \(\pi^3\), and MASt3R on \textbf{six} datasets,
\textbf{the same estimator on every backbone} lowers AUSE below built-in confidence in \textbf{15 of 18} conditions.
The staged evaluation yields \textbf{11 of 18} wins for the label-free core, $12/18$ for label-free
equal-weight fusion, $14/18$ with held-out weights, and $15/18$ when the built-in signal is included. Against a
trained evidential head, the result is a trade-off: the head calibrates magnitude better and leads in its
training domain, whereas \methodname{} transfers across backbones without adaptation. Its ranking improves point
filtering, but it does not detect stable systematic bias, aid novel-view synthesis, or transfer calibration across
domains.

\end{abstract}
\vspace{0.35cm}
  \end{@twocolumnfalse}
]

\section{Introduction}
\label{sec:intro}

Feed-forward 3D foundation models now recover dense scene geometry from a handful of images in one pass:
DUSt3R~\citep{dust3r}, MASt3R~\citep{mast3r}, VGGT~\citep{vggt}, and \pithree{}~\citep{pi3} regress per-pixel 3D
point maps without per-scene optimization or known camera poses. Those outputs carry no fixed coordinate frame,
so a far wall reconstructed correctly twice still appears unreliable: a fractional rotation between output frames
moves distant points enough to create large run-to-run variance. These predictions are useful only where reliability is established, yet their jointly
trained per-pixel confidence need not track geometric error. In the present evaluation, that correlation is weak in
distribution and weaker on average under shift, although unevenly so. The result limits point filtering, multi-view
fusion, and other decisions that depend on geometric reliability.

\paragraph{Perturbation uncertainty and its failure in 3D.}
Perturbation uncertainty offers a training-free alternative: rerun a frozen model on perturbed inputs and interpret
disagreement as uncertainty, as established for dense regression and depth~\citep{mi2022,poggi2020}. For 3D point
maps, that disagreement is not directly meaningful. Absolute scale, orientation, and position are unconstrained, so
the same geometry occupies different similarity gauges across runs. Native-frame variance then conflates
prediction instability with coordinate-frame variation. A closed form separates the desired
predictive spread from an additive, error-independent \emph{gauge} term that grows with squared distance from the
origin. In a large scene, this term potentially dominates and makes naive variance rank position rather than error.

\paragraph{Removing the gauge before the variance.}
The principle of removing a gauge before interpreting covariance is classical, but standard remedies act inside an
estimator through parameters, Jacobians, or normal equations. A frozen feed-forward network exposes only its
outputs. The available correction is correspondingly direct: align before variance. \methodname{} robustly
registers perturbed predictions to a common similarity (\Rthree{}) frame and then computes per-point variance,
without modifying or retraining the foundation model. Held-out calibration and rank-space fusion of perturbation
families are optional. The decomposition, simulation, and experiments on three architectures (VGGT, \pithree{},
MASt3R) across six datasets test this account. The recovered ranking depends comparatively little on the
perturbation family and improves multi-view point filtering; against the trained Trust3R head~\citep{trust3r}, it
presents a complementary trade-off rather than uniform superiority.

\paragraph{Contributions.}
\begin{enumerate}
\item \textbf{A closed form exposes gauge contamination in black-box perturbation ensembles.} Run-to-run
similarity-gauge variation adds an \emph{error-independent} term whose scale/rotation component grows with scene
extent. Classical remedies require access to the estimator, which the frozen-network setting does not provide.
Controlled simulation reproduces the effect, and all $30$ real VGGT view-sets tested exhibit its predicted
$\|x_p\|^2$ signature. The same confound arises whenever a predicted output carries a symmetry the model does not
observe.
\item \textbf{One output-side estimator applies across frozen point-map backbones.} Unlike a trained uncertainty
head tied to its checkpoint, \methodname{} applies without adaptation to any backbone that emits point maps. Robust
registration removes the nuisance without modifying or retraining the foundation model. A global \Rthree{}
correction directly instantiates the theory and accounts for $65\%$ of the empirical gain; the default then performs
evaluation-consistent per-view frame stabilization.
\item \textbf{Ablations isolate alignment as the source of the ranking gain.} Across VGGT, \pithree{}, and MASt3R on
six datasets, \methodname{} lowers AUSE below built-in confidence in \textbf{15 of 18} conditions, with the
supervision of each stage separated in Section~\ref{sec:main}. Ablations over alignment group, reference, scope,
aggregation, sample count, fit hyperparameters, and near-degenerate geometry isolate the gain, while the
camera-centre signature is strong on non-equivariant VGGT and near-absent on permutation-equivariant \pithree{}.
\item \textbf{The improved ranking has a defined use and boundary.} It improves multi-view point filtering, while
the comparison with a trained head remains a trade-off rather than a uniform ranking. It does not detect stable
systematic bias, transfer magnitude calibration across domains, or improve novel-view synthesis.
\end{enumerate}

\section{Related Work}
\label{sec:related}

\paragraph{Feed-forward 3D reconstruction.}
Direct point-map regression makes coordinate freedom an output-level concern. DUSt3R~\citep{dust3r} recasts
two-view geometry as per-pixel 3D points in a shared frame, and MASt3R~\citep{mast3r} adds metric-scale matching.
Many-view systems extend the setting: VGGT predicts point maps, depth, and cameras jointly~\citep{vggt}, and
\pithree{} is permutation-equivariant in the input views~\citep{pi3}. Thousand-image~\citep{fast3r},
streaming~\citep{spann3r,cut3r}, mixed-input~\citep{mapanything}, and any-view~\citep{depthanything3} systems
broaden the same family. Each system emits
geometry in a frame selected by the network, so the confound belongs to the setting rather than any one backbone. Each backbone's
\emph{global} point-map output is used (for \pithree{}, its world-frame \texttt{points}); view-order
equivariance does not remove that frame's similarity-gauge freedom. A second gap concerns reliability: the models'
per-pixel confidence channels are trained jointly with geometry rather than calibrated to geometric error, and are
frequently weak in the present evaluation, especially under distribution shift, consistent with concurrent VGGT evidence
on DTU~\citep{hillemann2026}.

\paragraph{Uncertainty estimation.}
Predictive uncertainty classically separates into aleatoric and epistemic components~\citep{kendall2017}.
Ensembles sample independently trained models~\citep{lakshminarayanan2017}; stochastic passes approximate that
sampling within one model~\mbox{\citep{gal2016}}; and deep evidential regression predicts distributional parameters in one
trained pass~\citep{amini2020}. All require training or model modification, and all risk degradation under dataset
shift~\citep{ovadia2019}. Input perturbations instead expose the sensitivity of a frozen
model~\citep{mi2022,wang2019tta}. This training-free strategy applies naturally to depth~\citep{poggi2020} and
flow~\citep{ilg2018}, where cross-run variance has a fixed coordinate meaning. Sparsification against an oracle
ordering measures ranking through AUSE~\citep{ilg2018}, while score-to-error scaling measures
calibration~\citep{guo2017,kuleshov2018}.
Point maps leave the unresolved case: global similarity freedom makes
their cross-run variance conflate instability with coordinate-frame variation.

\paragraph{Uncertainty for feed-forward 3D models.}
Trust3R~\citep{trust3r}, the closest method to the present setting, trains an evidential (Normal--Inverse--Wishart)
head~\citep{amini2020} on a frozen reconstruction backbone to predict per-point covariance in one pass from labelled
3D data. \methodname{} instead reads the unchanged checkpoint's response to equivalent inputs, exchanging additional
inference passes for zero adaptation. The methods answer the same reliability question under different access and
cost assumptions, which motivates a direct trade-off rather than a baseline to minimize.

\paragraph{Gauge freedom and alignment.}
Aligning point sets by a closed-form similarity fit is classical~\citep{umeyama1991}, and the deeper principle
behind it, that a free datum makes covariance uninterpretable, was established independently in geodetic free-network
adjustment~\citep{baarda1981}, statistical shape analysis~\citep{goodall1991}, bundle
adjustment~\citep{kanatani2001,morris2001,triggs2000}, and modern visual-inertial estimation~\citep{zhang2018};
this paper claims no novelty for it. These literatures also expose the missing case: every remedy acts on the
\emph{estimator}, through normal equations, inner constraints, or an optimizer covariance, and thus presumes
parameters and a Jacobian. A frozen feed-forward predictor provides none of them, only outputs.

Gauge normalization is routine at \emph{evaluation} time for point maps, which are scored after a similarity fit to
ground truth; Trust3R applies a per-image \Rthree{} alignment for this reason~\citep{trust3r}. Yet the gauge
\emph{inside} the ensemble remains untreated. Trust3R's released evaluator illustrates the split: it aligns each
image when scoring, while its ensemble and MC-dropout baselines accumulate cross-pass variance in raw predicted
frames. The missing output-side remedy is to register runs to a common frame before computing variance. The
resulting contamination is derived in closed form, and registration is tested for whether it separates an unusable
signal from a useful one.

\section{Problem: Perturbation Variance in Gauge-Ambiguous 3D}
\label{sec:problem}

The objective is to rank pixels by geometric error using only a frozen 3D model's outputs. Repeated-run variance has a clear
meaning in a fixed coordinate system; a point map is not constrained to a fixed frame across runs. This failure
is formalized first, then the gauge term the estimator must remove is derived.

\subsection{Setup and notation}
\label{sec:setup_notation}
Let the frozen feed-forward model $f$ map $V$ input views to per-view point maps
$X \in \mathbb{R}^{V\times H\times W\times 3}$. For pixel $p$ in view $v$, $X_{v}[p]\in\mathbb{R}^{3}$ is its
predicted position in the common output frame. The model $f$ is treated as a black box, and its weights remain unchanged.
Label-preserving perturbations (photometric jitter, view re-ordering, or small input noise) produce $N$ predictions
$\{X^{(n)}\}_{n=1}^{N}$ of the same scene. These predictions form a score $u[p]\ge 0$ that ranks pixels by geometric
error. Area Under the Sparsification Error (AUSE, lower is better) compares this ranking with an oracle
ordering~\citep{ilg2018}; Spearman correlation compares $u$ with realized error. That error is computed only after
gauge-free alignment to ground truth, so a global frame offset is not itself an error. Appendix~\ref{app:notation}
collects the notation.

The models also expose built-in confidence, but its correlation with error is weak in distribution and remains weak
under distribution shift (Section~\ref{sec:results}). This gap motivates a reliability signal derived from model
behaviour rather than a single confidence head.

\subsection{Why a direct variance is not well defined}
\label{sec:why_fails}
Perturbation methods read the spread of $\{X^{(n)}[p]\}_n$ as unreliability. This spread is meaningful for outputs
fixed to one coordinate system, including per-pixel labels, disparities, and depths. Images do not determine a point
map's absolute scale, orientation, or world position, so its geometry is defined only up to a global similarity
transform. Two runs therefore recover the same scene in different \emph{gauges}. Let run $n$ differ from a reference
by $T_n=(s_n, R_n, t_n)\in\Rthree{}$, with scale $s_n>0$, rotation $R_n\in SO(3)$, and translation
$t_n\in\mathbb{R}^{3}$. The native-frame variance $\mathrm{Var}_n[X^{(n)}[p]]$ then mixes predictive instability with
error-independent variation in $T_n$. Rotation and scale displace a point in proportion to its distance, so this
contamination grows with scene extent.

\subsection{Gauge contamination of the variance}
\label{sec:theory}
The effect is quantified with a small-perturbation gauge model.

\begin{assumption}[Small-perturbation gauge model]
\label{ass:gauge}
Let $x_p$ denote the clean geometry of point $p$. Each run adds isotropic predictive noise and applies the
similarity gauge $T_n=(s_n,R_n,t_n)$:
\begin{equation}
\begin{aligned}
X^{(n)}[p] &\;=\; s_n\,R_n\big(x_p+\delta_p^{(n)}\big)+t_n, \\
\delta_p^{(n)} &\sim \mathcal{N}\!\big(0,\ \sigma_{\mathrm{true}}^2(p)\,\mathbf{I}\big).
\end{aligned}
\label{eq:model}
\end{equation}
The gauge variables are small, mutually independent, independent of the predictive noise, and zero-mean about the identity:
\begin{equation}
\begin{aligned}
s_n &= 1+\varepsilon_n, \quad \varepsilon_n\sim\mathcal{N}(0,\sigma_s^2); \\
R_n &\approx \mathbf{I}+[\omega_n]_\times, \quad \omega_n\sim\mathcal{N}(0,\sigma_r^2\mathbf{I}); \\
t_n &\sim \mathcal{N}(0,\sigma_t^2\mathbf{I}).
\end{aligned}
\label{eq:gauge}
\end{equation}
\end{assumption}

\begin{proposition}[Gauge-contaminated variance]
\label{prop:decomp}
Under Assumption~\ref{ass:gauge} (Eqs.~\eqref{eq:model}--\eqref{eq:gauge}), to first order in the gauge parameters and for $\sigma_{\mathrm{true}}^2(p)\ll\|x_p\|^2$,
the per-point trace of the cross-run covariance separates additively as
\begin{equation}
U_{\mathrm{unaligned}}(p)\;=\;\underbrace{3\,\sigma_{\mathrm{true}}^2(p)}_{\text{true spread}}\;+\;
\underbrace{\big(\sigma_s^2+2\,\sigma_r^2\big)\,\|x_p\|^2 \;+\; 3\,\sigma_t^2}_{\text{gauge contamination}} .
\label{eq:decomp}
\end{equation}
\end{proposition}
Appendix~\ref{app:derivation} provides the derivation. The decomposition yields three consequences. First, the gauge
term is \emph{independent} of the target $\sigma_{\mathrm{true}}^2(p)$ and is a nuisance. Second, scale and rotation
contribute in proportion to the squared lever arm $\|x_p\|^2$; the factor two on $\sigma_r^2$ arises because a
fractional rotation moves a point in its orthogonal plane. For distant points or large scenes, this term potentially dominates
the true spread, making $U_{\mathrm{unaligned}}$ rank position rather than error. Third, translation adds the
position-independent offset $3\sigma_t^2$, which does not asymptotically reorder points. Centring each run removes
only this offset, not the persistent scale and rotation contamination. The variance becomes interpretable only after
the gauge is removed.

\paragraph{Scope of the gauge model.} Equation~\eqref{eq:decomp} describes one global similarity frame varying across
runs, not residual view-dependent frame motion in a multi-view predictor. Scene-level
alignment is accordingly evaluated; it instantiates Proposition~\ref{prop:decomp} and supplies most of the empirical gain, followed by the
default per-view refinement. Both estimator and target are defined modulo per-view alignment
(Section~\ref{sec:alignment}); this scope distinction determines the procedure that follows.

\section{Gauge-Aware Perturbation Uncertainty}
\label{sec:method}

\begin{figure*}[t]
\centering
\includegraphics[width=\textwidth]{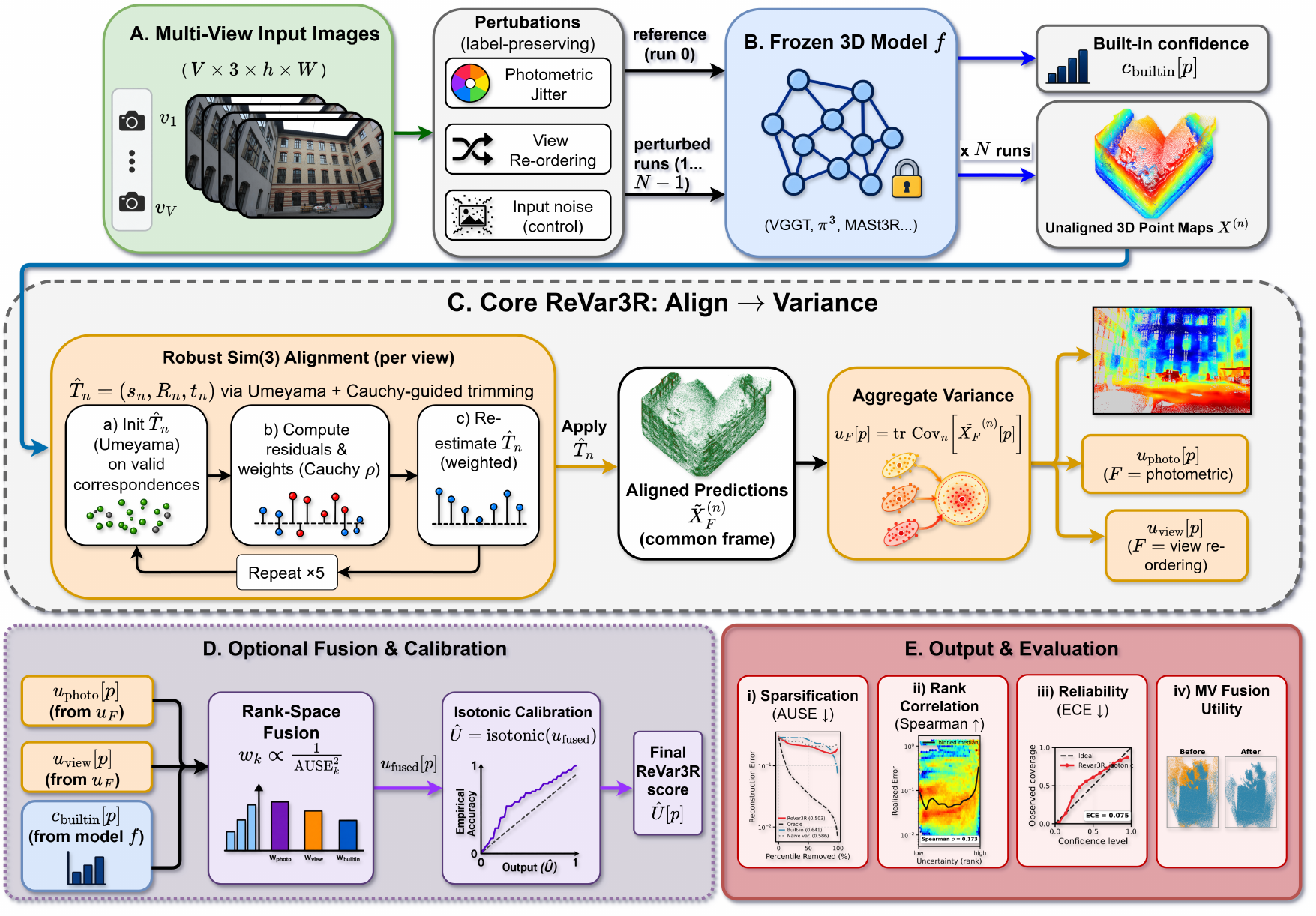}
\caption{\textbf{\methodname{} pipeline.} Alignment precedes per-point variance.}
\label{fig:hero}
\end{figure*}

\methodname{} targets the nuisance term in Proposition~\ref{prop:decomp} directly at the output level. Its procedure
has three stages: align each perturbed run to a shared frame, aggregate the aligned spread, then optionally calibrate
and fuse the scores. Figure~\ref{fig:hero} depicts the pipeline: aligning the perturbed predictions before taking
the variance localizes high scores on unstable geometry rather than on distant, correctly predicted points.
Algorithm~\ref{alg:revar3r} in Appendix~\ref{app:ablations} specifies the complete procedure. Output alignment is the
conceptual contribution; the
aggregation, calibration, and fusion operations are standard.

\subsection{Alignment before variance}
\label{sec:align}
\begin{definition}[Gauge-aligned variance]
\label{def:aligned}
For $N$ perturbed predictions of one view, run $0$ is selected as the reference, and each remaining run is mapped into
that frame by an estimated similarity transform $\hat{T}_n\in\Rthree{}$. The gauge-aligned variance of pixel $p$
is the trace of the per-point sample covariance of the \emph{aligned} predictions over the $N$ runs:
\begin{equation}
\tilde{X}^{(n)}[p]=\hat{T}_n\!\big(X^{(n)}[p]\big),\qquad
u[p]=\operatorname{tr}\,\mathrm{Cov}_n\!\big[\tilde{X}^{(n)}[p]\big].
\label{eq:aligned_var}
\end{equation}
\end{definition}
If $\hat{T}_n$ recovers each run's true gauge, alignment removes the additive term in Eq.~\eqref{eq:decomp} and
$u[p]\approx3\sigma_{\mathrm{true}}^2(p)$. Section~\ref{sec:results} shows that removing this nuisance before
aggregation recovers a useful error ranking, after which the perturbation family matters comparatively
little. Internal registration is strictly prediction-to-prediction. It uses prediction-derived correspondences only;
ground-truth validity is used exclusively for evaluation metrics. At evaluation time, a separate gauge-free
alignment defines prediction error as described in Section~\ref{sec:setup}.

\subsection{Robust similarity alignment}
\label{sec:robust}
$\hat{T}_n$ is initialized with a closed-form similarity (\Rthree{}) fit~\citep{umeyama1991}, then refined through
five rounds of robust trimming under a Cauchy $M$-estimator scale~\citep{zhang1997}. Plain least squares would allow
the unstable or high-error regions being \emph{detected} to pull the gauge estimate toward themselves and partially
reabsorb their signal. In each round, Cauchy weights rank the correspondences, the worst $10\%$ are discarded, and
the closed-form fit is recomputed on the remainder. The weights determine rejection order; they do not enter the fit
as continuous weights. This procedure anchors registration to stable geometry, so the residual variance reflects
instability rather than fit error.

For each run, one $\hat{T}_n$ is estimated relative to the reference from all finite, prediction-valid correspondences. No
confidence threshold or manual mask is applied. The seven parameters are over-determined well below a view's full point
count, so the fit uses a fixed-seed random subsample of at most $8{,}000$ correspondences and then applies it to
every point. At the point counts evaluated here, it is insensitive to trimming fraction, Cauchy scale,
and near-planar geometry (Appendix~\ref{app:robustalign}). The same appendix controls explicitly for the exclusion
of ground-truth validity from this internal fit.

\subsection{Aggregation, calibration, and fusion}
\label{sec:aggfuse}
The core estimator aggregates aligned spread with the trace of the per-point covariance
(Eq.~\eqref{eq:aligned_var}). Trace variance produces a ranking without ground truth; robust median spread is weaker
in the ablation, while the largest-eigenvalue aggregator is marginally weaker
(Appendix~\ref{app:ablations}). Two optional components use a small held-out labelled split without changing any
single family's ranking. Isotonic calibration~\citep{kuleshov2018} maps a score to error magnitude for expected
calibration error (ECE). Rank-space fusion combines photometric jitter, view re-ordering, and built-in confidence,
weighting each signal by held-out inverse-AUSE$^2$ so that a weak family is down-weighted rather than allowed to
degrade the estimate. Aligned single-family variance is termed the core, and the fused variant is termed the full \methodname{}
estimator. Table~\ref{tab:levels} separates the label-free ranking from held-out fusion and magnitude calibration.

\section{Experimental Setup}
\label{sec:setup}

\paragraph{Models and datasets.}
The evaluation uses released checkpoints for three frozen backbones: VGGT~\citep{vggt}, \pithree{}~\citep{pi3}, and
MASt3R~\citep{mast3r}. VGGT and MASt3R are reference-dependent; \pithree{} is permutation-equivariant. The six
datasets span complementary regimes. 7-Scenes~\citep{shotton2013} and DTU~\citep{jensen2014} are in-distribution.
ETH3D~\citep{schoeps2017} is a wide-baseline outdoor shift, while BlendedMVS~\citep{yao2020} overlaps the base
models' training corpus, where built-in confidence is already strong. The competitor's own benchmarks
(Section~\ref{sec:trust3r}) add indoor RGB-D TUM~\citep{sturm2012} and outdoor-driving KITTI~\citep{geiger2012};
KITTI's sparse LiDAR ground truth and road scenes are far from object- and room-scale training data. Trust3R also
reports ScanNet++, which is unavailable here, so the evaluation covers three of its four benchmarks (TUM, KITTI, ETH3D). Each
condition contains overlapping $V{=}5$-view sets with per-pixel ground-truth point maps. Ground truth is resized to
the model output resolution, and comparisons follow gauge-free alignment.

\paragraph{Metrics.}
The primary metric, Area Under the Sparsification Error (AUSE)~\citep{ilg2018} (\down{}), compares an uncertainty
ordering with the oracle error ordering after per-scene normalization. The evaluation also reports Spearman correlation (\up{})
with realized error as a complementary monotone-invariant ranking measure, Area Under the Random Gain (AURG,
\up{}) as a reference, and regression ECE (\down{}) for magnitude calibration~\citep{kuleshov2018}. Per-point error
is Euclidean distance to ground truth after robust similarity alignment. This evaluation alignment is separate from
the estimator's internal run-to-run registration (Section~\ref{sec:method}); global frame offsets are not scored as
error.

\paragraph{Baselines.}
Training-free baselines comprise built-in confidence, an input-noise (Monte Carlo, MC) ensemble taken both
without alignment and through the identical registration and scoring path as \methodname{}, and random and oracle
references. The aligned ensemble is the stronger comparator and is reported as such
(Table~\ref{tab:alignedmc}). The same-task learned comparator, Trust3R~\citep{trust3r}, trains an
evidential uncertainty head on frozen MASt3R. Its released checkpoint operates at $224$~px, so the head-to-head
evaluation uses that backbone at matched resolution and on identical point sets (Section~\ref{sec:trust3r}).

\paragraph{Protocol and statistics.}
Unless noted, \methodname{} uses $N{=}5$ passes for each of two families: random view re-ordering and photometric
jitter, with brightness, contrast, and colour each scaled by ${\pm}25\%$ plus ${\approx}0.5\%$ additive pixel noise.
The families share their reference prediction, so they require 9 total model evaluations. The naive input-noise
baseline adds ${\approx}1.5\%$ Gaussian pixel noise. Appendix~\ref{app:nsweep} motivates the sample count.

The view-set, not the pixel, is the unit of statistical inference because spatially correlated per-pixel errors would
overstate significance. AUSE is computed per view-set, paired differences over view-sets are tested with the Wilcoxon
signed-rank test, and report absolute and relative effect sizes. View-sets from one scene are also correlated, making
this test anticonservative when a dataset supplies few independent scenes. For the trained-head comparison, effect sizes are therefore
reported rather than $p$-values except where the scene count supports a test
(Appendix~\ref{app:reconcile}). Win counts such as $15/18$ are descriptive and carry no test.

Within each backbone, the training-free estimators share predictions, alignment conventions, and scoring
code; Trust3R uses its own refined point map, scored on the same valid pixels. The stochastic estimators
(\methodname{} and the input-noise ensemble) run over five random seeds, and the seed mean is reported; the trained
Trust3R head and built-in confidences are deterministic single-pass references (Appendix~\ref{app:seeds}). In the
Trust3R comparison (\S\ref{sec:trust3r}), both methods use identical view-sets
(all available view-sets per dataset; counts in Table~\ref{tab:trust3r}), so per-point AUSE covers the same points.

\section{Results}
\label{sec:results}

The experiments follow the causal chain of the argument: isolate gauge contamination, remove it on real models,
then compare the recovered ranking with built-in confidence, under shift, against a trained head, and in point
filtering.

\subsection{Controlled validation of gauge contamination}
\label{sec:synthetic}
\begin{figure}[!b]
\centering
\includegraphics[width=0.82\linewidth]{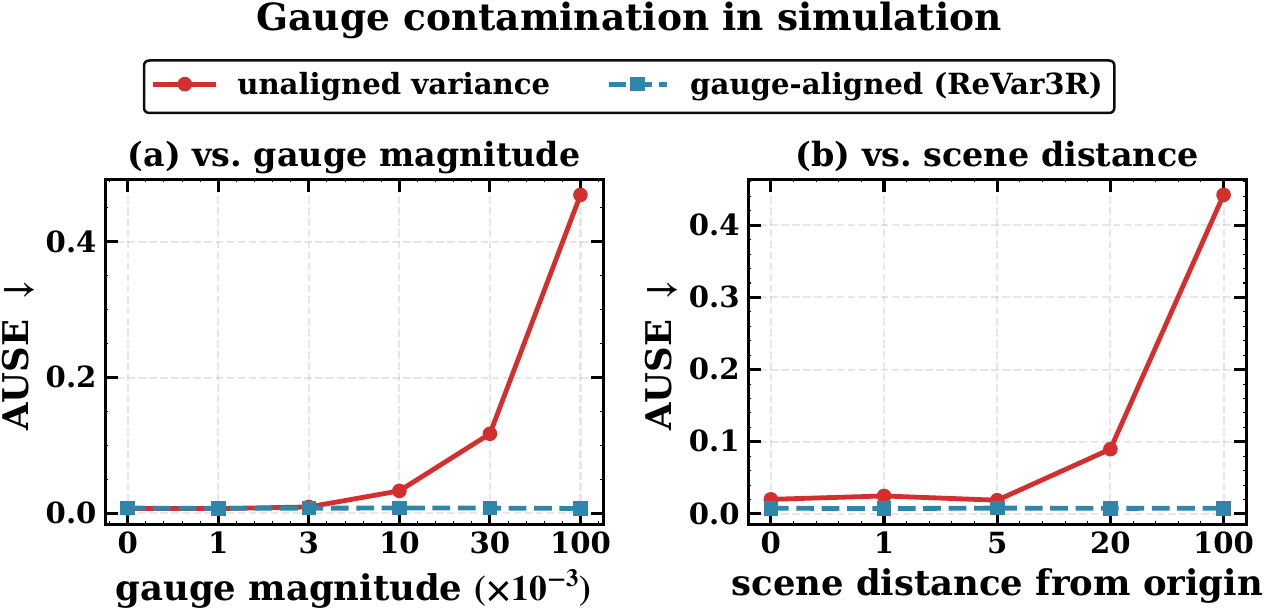}
\caption{\textbf{Gauge contamination in simulation.} (a) A growing injected \Rthree{} gauge degrades unaligned
perturbation variance (AUSE~\down{}); gauge-aligned variance stays near oracle quality. (b) At a fixed gauge, the
degradation grows with the scene's distance from the origin, the lever arm in the $\|x\|^2$ term of
Eq.~\eqref{eq:decomp}.}
\label{fig:synthetic}
\end{figure}

Alignment removes injected gauge contamination; unaligned variance deteriorates as the gauge grows. The validation uses known
per-point uncertainty $\sigma_{\mathrm{true}}$, a controlled \Rthree{} gauge, and the real-data estimator code
(Appendix~\ref{app:derivation}). Increasing the gauge degrades unaligned ranking
(Spearman $\mathbf{0.99\to0.40}$; AUSE $\mathbf{0.007\to0.47}$), while aligned variance remains at oracle
quality (Spearman $\mathbf{\approx0.99}$, AUSE $\mathbf{\approx0.008}$) across both gauge magnitude and scene
distance (Figure~\ref{fig:synthetic}). The transform decomposition matches Eq.~\eqref{eq:decomp}: scale and rotation
produce persistent distance-scaling contamination, whereas translation contributes only a position-independent
offset.

Real models reproduce that signature. For every one of $30$ VGGT view-sets across six datasets,
$U_{\mathrm{unaligned}}-U_{\mathrm{aligned}}$ grows with $\|x_p\|^2$ (median Spearman $\mathbf{+0.59}$,
positive in \textbf{$30/30$} view-sets; Appendix~\ref{app:derivation}). A direct fit of Eq.~\eqref{eq:decomp} across
all three backbones and six datasets yields a positive gauge coefficient in $\mathbf{16/18}$ conditions and median
$R^2=\mathbf{0.74}$. The predicted functional form, not only its direction, holds on real models
(Appendix~\ref{app:derivation}).

\subsection{Alignment removes the dominant nuisance}
\label{sec:alignment}
Among the estimator's stages, alignment rather than fusion with the baseline produces the main ranking gain.
Averaged over six datasets, it
lowers mean AUSE by $\mathbf{0.165}$, $\mathbf{0.131}$, and $\mathbf{0.105}$ for VGGT, MASt3R, and \pithree{}
(Figure~\ref{fig:align}). The reduction is larger in distribution ($+0.167$ mean, against $+0.066$ under shift), and wide-baseline
ETH3D is the only dataset with near-zero mean alignment benefit across backbones. Robust alignment is at least as good
as a plain similarity fit on every backbone (Appendix~\ref{app:ablations}). The \emph{training-free core}, aligned
single-family variance without fusion or built-in confidence, beats the built-in signal in $\mathbf{11/18}$
conditions, versus $5/18$ unaligned (Tables~\ref{tab:coredecomp},~\ref{tab:newds}). Held-out inverse-AUSE$^2$ fusion
adds less, beating naive equal fusion in $83\%$ of conditions (Appendix~\ref{app:ablations}).

Two ablations identify the nuisance removed (Appendix~\ref{app:alignabl}). \textbf{(i) The group matches the theory:}
AUSE falls from none $0.543$, through translation $0.500$ and rigid $\mathrm{SE}(3)$ $0.460$, to similarity
$\mathrm{Sim}(3)$ $\mathbf{0.392}$. The largest steps add rotation and scale, removing the
$(\sigma_s^2+2\sigma_r^2)\|x\|^2$ nuisance in Eq.~\eqref{eq:decomp}. \textbf{(ii) The signal belongs to the ensemble:}
fixed run-$0$, medoid, and random-run references give indistinguishable AUSE ($0.392/0.395/0.394$), requiring no
privileged anchor.

\paragraph{Global gauge removal explains most, but not all, of the alignment gain.}
One scene-level transform, the run-level nuisance of Proposition~\ref{prop:decomp}, lowers mean
AUSE $0.543\to0.446$, accounting for $0.097$ of the default alignment's $0.151$ total gain, or $65\%$; per-view
transforms supply the remaining $35\%$ ($0.446\to\mathbf{0.392}$, helping $11/18$ vs.\ $7/18$), which
Appendix~\ref{app:alignabl} interprets. Routing an input-noise ensemble through the same registration lifts it
from beating built-in confidence in five of nine conditions to all nine (Table~\ref{tab:alignedmc}).

\subsection{Comparison with built-in confidence}
\label{sec:main}
The same estimator improves over built-in confidence in most conditions, but not entirely without labels. \methodname{} lowers AUSE in \textbf{15 of 18} conditions across three backbones $\times$ six datasets. Its
four-stage ladder exposes supervision: the label-free core wins $\mathbf{11/18}$; label-free
equal-weight fusion reaches $12/18$; fitting fusion weights on a held-out split reaches $14/18$; and admitting the
built-in signal as a fusion input reaches $15/18$ (Appendix~\ref{app:ablations}). Only the first two stages use no
labels at any point. The core supplies most of the effect; reusing the baseline recovers one condition. Repeating
the full grid over five seeds returns $15/18$ every time, with no condition changing sign
(Table~\ref{tab:seedgrid}).

The wins include all six TUM and KITTI conditions at each backbone's native resolution
(Figure~\ref{fig:mainheat}, positive cells; exact values in Appendices~\ref{app:perscene} and~\ref{app:newds}). The three
exceptions are \pithree{} and MASt3R on BlendedMVS, which overlaps the base models' training corpus, and \pithree{}
on ETH3D. Built-in confidence is comparatively strong in these cells, although not uniformly strongest in the grid
(Appendix~\ref{app:perscene}). Gains are largest where that signal is weak: VGGT on 7-Scenes
($0.712\to0.491$, $-31\%$), MASt3R on 7-Scenes ($0.587\to0.358$, $-39\%$), and VGGT on KITTI
($0.401\to0.237$, $\mathbf{-41\%}$). Appendix~\ref{app:qual} shows representative uncertainty maps for
7-Scenes, DTU, and ETH3D, with the typical gains and the variability against built-in confidence.

\begin{figure*}[t]
\centering
\includegraphics[width=\textwidth]{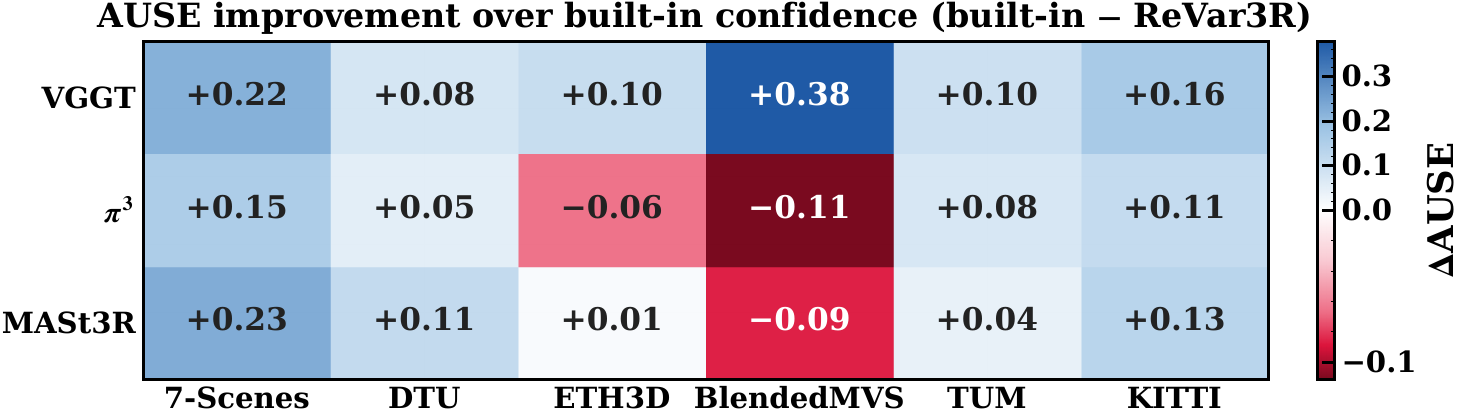}
\caption{\textbf{Main comparison over six datasets.} AUSE improvement (built-in $-$ \methodname{}) per backbone and
dataset; blue is a gain and red a loss, darker is larger. Positive cells (\textbf{15 of 18}) are conditions where \methodname{} lowers AUSE.}
\label{fig:mainheat}
\end{figure*}

\subsection{Behaviour under distribution shift}
\label{sec:ood}
The gain under shift is not uniform. On far-shift outdoor KITTI, \methodname{} ranks error best (Spearman
$\mathbf{+0.69}$ against $+0.48$ for built-in confidence and $+0.44$ for the trained head) and attains the lowest
AUSE ($0.216$ against $0.359$). ETH3D, by contrast, is a loss: across $24$ wide-baseline view-sets spanning nine
scenes, built-in confidence ranks better than \methodname{} ($+0.31$ against $+0.13$) and has lower
AUSE ($0.585$ against $0.667$); the trained head is better still (Appendix~\ref{app:ood}). Alignment helps most where the frame moves and the baseline is weak (Section~\ref{sec:alignment}).

\subsection{Relation to a trained uncertainty head}
\label{sec:trust3r}
The trained-head comparison exposes a trade-off, not a uniform ranking. Trust3R~\citep{trust3r} attaches an
evidential head to frozen MASt3R; its released checkpoint enables a matched $224$~px comparison on identical point
sets across five datasets, including TUM and KITTI (Table~\ref{tab:trust3r}). By per-view-set median AUSE,
\methodname{} is favored on three of five datasets; the trained head leads on TUM, and 7-Scenes is a near tie
($0.272$ against $0.271$). By Spearman, the trained head leads on two, and it calibrates magnitude better throughout.

Only DTU supports scene-level significance, because view-sets from one scene are correlated; sampling one view-set
from each of $30$ DTU scans, \methodname{} has lower AUSE on \textbf{every scan}; the per-scene mean difference is $\mathbf{-0.205}$,
and the scene-level sign-flip gives $p=1.9\times10^{-9}$ (Appendix~\ref{app:reconcile}). \methodname{} adds no
parameters but costs ${\sim}10\times$ a single-pass head (Appendix~\ref{app:cost}). The trained head is stronger and
better calibrated in its training domain; the training-free estimator holds up on the most dissimilar data.

\subsection{Downstream utility}
\label{sec:downstream}
The improved ranking retains better geometry after uncertain points are removed. Under the selective-prediction
protocol~\citep{geifman2017}, the most-uncertain points are removed, and mean error is measured among those kept
(Figure~\ref{fig:downstream}). At $50\%$ removal, \methodname{} closes $\mathbf{43\%}$ on 7-Scenes,
$\mathbf{57\%}$ on DTU, and $\mathbf{30\%}$ on outdoor KITTI of the built-in-to-oracle gap. It beats built-in,
input-noise, and random removal on every dataset.

\section{Discussion and Limitations}
\label{sec:discussion}

\paragraph{Limitations.}
Four limitations follow. First, \methodname{} measures \emph{instability}, not \emph{incorrectness}: stable
systematic error has low variance when perturbations agree on the same wrong geometry, a limitation the trained head
also retains (Appendix~\ref{app:negatives}). Second, magnitude calibration requires labels and transfers poorly
across domains for both methods. An exchangeable in-domain calibration set could support
finite-sample marginal coverage guarantees through conformal prediction~\citep{angelopoulos2023}.
Third, uncertainty-guided initialization does not improve novel-view synthesis
under a fixed budget. Fourth, multiple passes add roughly an order of magnitude in latency
(Appendix~\ref{app:cost}), and the analysis idealizes the estimated gauge (Appendix~\ref{app:scope}).

\paragraph{Broader implication.}
Normalize an unobserved symmetry before interpreting variance in a structured geometric output. Camera poses confirm
the mechanism in a second modality because they share the point-map frame: naive centre uncertainty has the predicted
$\|c\|^2$ inflation, strong for
non-equivariant VGGT ($p<10^{-12}$ on all three datasets) and near-absent on permutation-equivariant \pithree{}
outside 7-Scenes. Alignment improves centre AUSE in all six conditions, but centre-error Spearman confirms only three
of six, leaving pose instability a weak error ranker in some settings (Appendix~\ref{app:poses}). The principle should extend to shape codes under global similarity and to radiance
fields (NeRF~\citep{mildenhall2020}, 3D Gaussian Splatting~\citep{kerbl2023}), whose geometry is defined only up to
a world frame~\citep{goli2024}.

\section{Conclusion}
\label{sec:conclusion}

Perturbation uncertainty for feed-forward 3D point maps becomes interpretable only after similarity-gauge
variation is removed. The derived additive term and simulation show that alignment recovers a near-oracle signal.
\methodname{} applies align-before-variance to frozen-model outputs without added parameters or retraining,
improving ranking over built-in confidence in most conditions and in point filtering, though not uniformly under
shift. It measures gauge-aware procedural reliability, not geometric error: it misses stable systematic bias, and
its calibration does not transfer across domains.

\subsection*{Reproducibility statement}
\label{app:repro}
Section~\ref{sec:method} specifies the estimator: the two perturbation families, robust similarity
registration with its trimming schedule and fit subsample, trace aggregation, held-out isotonic
calibration, and inverse-AUSE$^2$ rank fusion. Section~\ref{sec:setup} lists the public backbone and
Trust3R checkpoints, the six datasets, the perturbation settings, and the statistical protocol,
including per-view-set paired Wilcoxon tests and the five-seed convention.
Appendix~\ref{app:derivation} gives the full derivation of the gauge term and its Monte-Carlo check,
Appendix~\ref{app:reconcile} the identical-point-set protocol of the Trust3R comparison, and
Appendix~\ref{app:seeds} the seed-stability analysis. Appendices~\ref{app:perscene}
and~\ref{app:newds} report the per-condition values behind the aggregates in the main text.

\subsection*{AI use statement}

Large language models are used in the preparation of this paper for two purposes, both under author
supervision and both disclosed in the submission form. First, to draft sections of the paper: passages are
generated from author-supplied outlines, derivations, and result tables, then edited by the authors. Second, to
aid or polish the writing: previously drafted prose is revised for concision, consistency of terminology, and
clarity. Every drafted or polished passage is read, corrected, and approved by the authors before inclusion.

The research problem, conceptual framework, hypotheses, theoretical results, and benchmark design are developed
by the authors before any generative AI is used. AI is not used to originate ideas or hypotheses, to retrieve or
select related work, to derive or verify proofs, or to produce any reported result. Every citation is checked
against its original source, and every reported number comes from the reported experiments. The authors take
full responsibility for the manuscript and all claims, results, and artifacts.

\bibliography{refs}
\bibliographystyle{iclr2027_conference}

\clearpage
\appendix
\section*{Supplementary Material}
\emph{\methodname{}: Gauge-Aware Perturbation Uncertainty for Feed-Forward 3D Reconstruction}
\section{Notation}
\label{app:notation}

This section fixes the notation shared by the problem statement (Section~\ref{sec:problem}), estimator
(Section~\ref{sec:method}), and derivation (Appendix~\ref{app:derivation}). Table~\ref{tab:notation} groups the
symbols by their role in the model, gauge, estimator, and evaluation.

\begin{table*}[b]
\caption{\textbf{Notation.} Definitions used throughout the paper, grouped by model, perturbation ensemble, gauge,
estimator, and metric.}
\label{tab:notation}
\begin{center}\begin{tabular}{ll}
\toprule
\textbf{Symbol} & \textbf{Meaning} \\
\midrule
\multicolumn{2}{l}{\emph{Model and geometry}} \\
$f$ & frozen feed-forward 3D model (weights never modified) \\
$V$ & number of input views \\
$H,\,W$ & height and width of each per-view point map \\
$X\in\mathbb{R}^{V\times H\times W\times 3}$ & per-view point maps emitted by $f$ \\
$p,\;v$ & pixel (correspondence) index; view index \\
$X_v[p]\in\mathbb{R}^{3}$ & predicted 3D position of pixel $p$ in view $v$ \\
$x_p\in\mathbb{R}^{3}$ & clean (true) geometry of point $p$ \\
$\|x_p\|$ & distance of $x_p$ from the frame origin (lever arm) \\
$\delta_p^{(n)}$ & isotropic predictive noise, $\mathcal{N}(0,\sigma_{\mathrm{true}}^2(p)\mathbf{I})$ \\
$\sigma_{\mathrm{true}}^2(p)$ & true per-point predictive variance (the target signal) \\
\midrule
\multicolumn{2}{l}{\emph{Perturbation ensemble}} \\
$N$ & number of perturbed passes per family \\
$X^{(0)}$ & reference prediction (run $0$) \\
$\{X^{(n)}\}_{n=1}^{N}$ & the $N$ perturbed predictions; $n$ is the run index \\
$\mathcal{F},\;F$ & set of perturbation families; a single family \\
$P_n^{F}$ & the $n$-th label-preserving input perturbation drawn from $F$ \\
\midrule
\multicolumn{2}{l}{\emph{Gauge (similarity transform)}} \\
$\Rthree{}$ & similarity group ($7$ DoF: scale, rotation, translation) \\
$SO(3)$ & rotation group \\
$T_n=(s_n,R_n,t_n)$ & true per-run similarity gauge, $T_n\in\Rthree{}$ \\
$\hat{T}_n$ & fitted similarity map from run $n$'s frame to the reference frame \\
$s_n=1+\varepsilon_n$ & per-run scale; $\varepsilon_n\sim\mathcal{N}(0,\sigma_s^2)$ \\
$R_n\approx\mathbf{I}+[\omega_n]_\times$ & per-run rotation; $\omega_n\sim\mathcal{N}(0,\sigma_r^2\mathbf{I})$ \\
$[\omega_n]_\times$ & skew-symmetric (cross-product) matrix of $\omega_n$ \\
$t_n$ & per-run translation, $\mathcal{N}(0,\sigma_t^2\mathbf{I})$ \\
$\sigma_s^2,\,\sigma_r^2,\,\sigma_t^2$ & scale, rotation, translation gauge-jitter variances \\
\midrule
\multicolumn{2}{l}{\emph{Estimator}} \\
$\tilde{X}^{(n)}[p]=\hat{T}_n(X^{(n)}[p])$ & gauge-aligned prediction \\
$\mathrm{Cov}_n[\cdot],\;\operatorname{tr}$ & per-point sample covariance across runs; its trace \\
$u[p]\ge 0$ & per-point uncertainty score (Eq.~\eqref{eq:aligned_var}) \\
$U_{\mathrm{unaligned}}(p)$ & naive unaligned trace-covariance (Eq.~\eqref{eq:decomp}) \\
$u_F[p]$ & per-family uncertainty before fusion \\
$\rho$ & Cauchy $M$-estimator (robust) loss \\
\midrule
\multicolumn{2}{l}{\emph{Metrics}} \\
AUSE & Area Under the Sparsification Error (lower is better) \\
Spearman & rank correlation of $u$ with realized error \\
AURG & Area Under the Random Gain (higher is better) \\
ECE & Expected Calibration Error (lower is better) \\
\bottomrule
\end{tabular}
\end{center}
\end{table*}

\clearpage
\section{Gauge Mechanism and Verification}
\label{app:derivation}

This section establishes the closed-form gauge-contamination term, verifies it numerically, and tests its predicted
distance signature in simulation and real scenes.

\paragraph{Derivation of Eq.~\eqref{eq:decomp}.}
Under Eqs.~\eqref{eq:model}--\eqref{eq:gauge}, each run is linearized about the identity gauge. Substituting
$R_n\approx\mathbf{I}+[\omega_n]_\times$ and $s_n=1+\varepsilon_n$, then retaining first-order terms in
$(\varepsilon_n,\omega_n,t_n)$, gives
\begin{equation}
X^{(n)}[p]-x_p \;\approx\; \delta_p^{(n)} \;+\; \varepsilon_n\,x_p \;+\; \omega_n\times x_p \;+\; t_n,
\label{eq:expand}
\end{equation}
where $\delta_p^{(n)}$ is predictive noise and $(\varepsilon_n,\omega_n,t_n)$ are the scale, rotation, and
translation parameters defined in Section~\ref{sec:theory}. Mutual independence and zero means eliminate the
cross-covariances, so the per-point covariance is the sum
\begin{align*}
\mathrm{Cov}\big(\delta_p^{(n)}\big) &= \sigma_{\mathrm{true}}^2(p)\,\mathbf{I}, &
\mathrm{Cov}\big(t_n\big) &= \sigma_t^2\,\mathbf{I}, \\
\mathrm{Cov}\big(\varepsilon_n\,x_p\big) &= \sigma_s^2\,x_px_p^\top, &
\mathrm{Cov}\big(\omega_n\times x_p\big) &= \sigma_r^2\big(\|x_p\|^2\,\mathbf{I}-x_px_p^\top\big),
\end{align*}
where $\mathbf{I}$ is the $3\times3$ identity. The trace is then taken. Because
$\operatorname{tr}(\mathbf{I})=3$ and $\operatorname{tr}(x_px_p^\top)=\|x_p\|^2$, the rotation covariance contributes
$\sigma_r^2(3-1)\|x_p\|^2=2\sigma_r^2\|x_p\|^2$. The two isotropic terms therefore sum to
$3\sigma_{\mathrm{true}}^2(p)+3\sigma_t^2$, while the scale and rotation terms sum to
$(\sigma_s^2+2\sigma_r^2)\|x_p\|^2$, yielding Eq.~\eqref{eq:decomp}. The first-order expansion also generates the
cross terms $\varepsilon_n\delta_p^{(n)}$ and $\omega_n\times\delta_p^{(n)}$. Their contribution,
$3\sigma_{\mathrm{true}}^2(\sigma_s^2+2\sigma_r^2)$, only rescales the signal by
$(1+\sigma_s^2+2\sigma_r^2)$ and cannot reorder points. These terms are therefore omitted under
$\sigma_{\mathrm{true}}^2\ll\|x_p\|^2$. Trace variance is reported; the estimator uses its rank-equivalent square root.

\paragraph{Synthetic experiment.} Alignment removes the injected gauge effect, whereas unaligned ranking degrades
with gauge magnitude and scene distance. The experiment draws heterogeneous known $\sigma_{\mathrm{true}}$, injects a known
\Rthree{} gauge per run, and uses the same align/aggregate implementation as for real data. Gauge magnitude $g$ is
the standard deviation of the scale factor and of the rotation angle (in radians), and translation has standard
deviation $5g$ per axis. The points fill a cube of side $2$ centred at distance $5$ from the origin; the distance
sweep moves this cube at $g=0.01$. Table~\ref{tab:synth}
also isolates components and sample count: scale$+$rotation creates persistent $\|x\|^2$-scaling contamination,
whereas translation adds a position-independent offset that does not reorder points (Eq.~\eqref{eq:decomp}). The
$N$-sweep shows that the scale$+$rotation contamination persists instead of averaging out with $N$.

\begin{table*}[t]
\caption{\textbf{Synthetic gauge contamination.} Spearman against true uncertainty and AUSE show that aligned
estimates remain invariant while unaligned estimates degrade with gauge magnitude and scene distance. Component
rows use magnitude $0.05$; the $N$-sweep reports unaligned Spearman and shows persistent contamination.}
\label{tab:synth}
\begin{center}\begin{tabular}{llcccc}
\toprule
\textbf{Sweep} & \textbf{setting} & \textbf{$\pmb{\rho_{\text{unaligned}}}$} & \textbf{$\pmb{\rho_{\text{aligned}}}$} & \textbf{AUSE$_{\text{un}}$} & \textbf{AUSE$_{\text{al}}$} \\
\midrule
gauge mag. & $0.0$ & 0.991 & 0.990 & 0.007 & 0.008 \\
 & $0.01$ & 0.936 & 0.989 & 0.033 & 0.008 \\
 & $0.03$ & 0.774 & 0.989 & 0.117 & 0.008 \\
 & $0.10$ & 0.395 & 0.990 & 0.469 & 0.007 \\
\midrule
distance & $D{=}0$ & 0.962 & 0.989 & 0.020 & 0.008 \\
 & $D{=}20$ & 0.852 & 0.989 & 0.090 & 0.008 \\
 & $D{=}100$ & 0.477 & 0.989 & 0.442 & 0.008 \\
\midrule
component & translation only & 0.808 & $-$ & $-$ & $-$ \\
 & scale$+$rotation & 0.679 & $-$ & $-$ & $-$ \\
$N$-sweep & scale$+$rot ($N{=}8/50/200$) & 0.74/0.84/0.80 & $-$ & $-$ & $-$ \\
\bottomrule
\end{tabular}
\end{center}
\end{table*}

\begin{figure}[htb]
\centering
\includegraphics[width=0.66\linewidth]{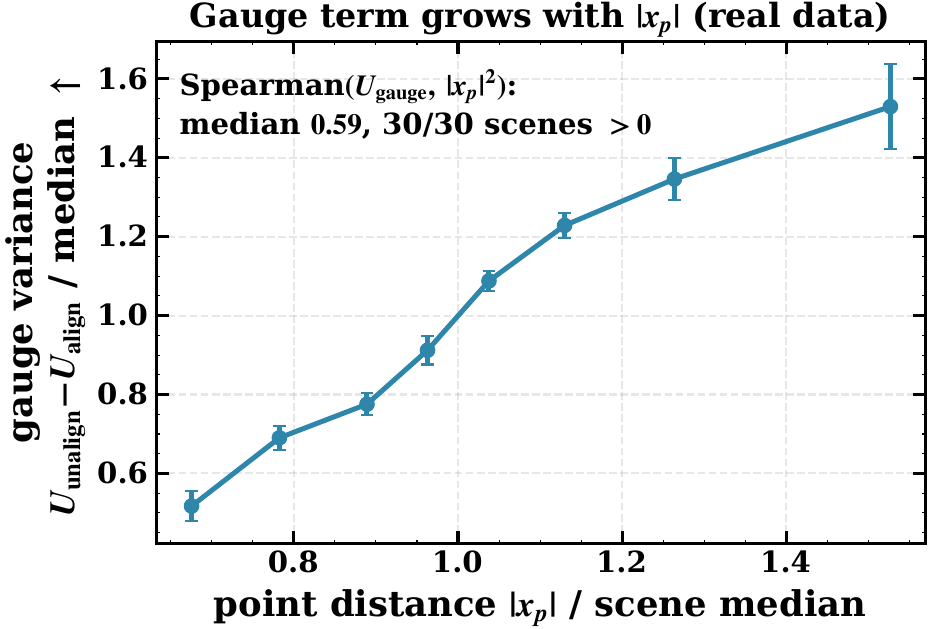}
\caption{\textbf{The gauge term grows with distance within real scenes.} After per-view-set normalization and
pooling over $30$ view-sets from six datasets (eight scenes in all),
$U_{\mathrm{unaligned}}-U_{\mathrm{aligned}}$ rises monotonically with point distance $|x_p|$. Its rank correlation
with $|x_p|^2$ is positive in $30/30$ view-sets (median $+0.59$), matching the $\|x_p\|^2$ term in
Eq.~\eqref{eq:decomp}.}
\label{fig:xnorm}
\end{figure}

\paragraph{Real-data confirmation of the $\|x_p\|^2$ term.} Every tested view-set exhibits the predicted
within-view-set distance signature. Unlike the simulation, real data do not control gauge magnitude, but the
theory's \emph{shape} prediction remains testable \emph{within} each view-set. For each of $30$ view-sets ($5$ per
dataset $\times$ six datasets, VGGT; those five come from one or two scenes per dataset, eight scenes in all, so
these are repeated views of a scene rather than thirty independent scenes), the gauge-attributable variance is computed per point as
$U_{\mathrm{unaligned}}(p)-U_{\mathrm{aligned}}(p)$, and its rank correlation with $\|x_p\|^2$ (point
distance in the model's output frame, where the gauge acts). It is positive in \textbf{all $30$} (median
Spearman $\mathbf{+0.59}$; per-dataset medians $+0.53$ to $+0.88$), and the pooled decile-binned curve rises
monotonically with distance (Figure~\ref{fig:xnorm}), confirming the distance-growing gauge term of
Eq.~\eqref{eq:decomp} on real models. A clean \emph{cross-dataset} extent trend does not hold because uncontrolled
per-scene gauge magnitude confounds it. The within-scene test, rather than a cross-dataset comparison, is what
isolates the $\|x_p\|^2$ shape from that magnitude.

\paragraph{Recovering the gauge coefficient across all eighteen conditions.} The rank test above establishes the
direction of the effect; the functional form permits a direct test. For each of the $3$ backbones $\times$ $6$
datasets, the analysis records per-point $U_{\mathrm{unaligned}}$ and $U_{\mathrm{aligned}}$ from the same forward passes, bins
points into $50$ quantiles of $\|x_p\|^2$, and fits the conditional means to Eq.~\eqref{eq:decomp}, whose slope is
the gauge coefficient $(\sigma_s^2+2\sigma_r^2)$ and whose intercept is $3\sigma_t^2$. Binning is necessary because
a per-point variance from $N{=}5$ samples carries ${\approx}70\%$ relative noise, which bounds any per-point fit
regardless of whether the model is correct; Eq.~\eqref{eq:decomp} is a statement about the conditional expectation.
The affine form fits well: median $R^2=\mathbf{0.74}$ over the eighteen conditions (max $0.995$), with a positive
recovered coefficient in $\mathbf{16/18}$. The two exceptions (VGGT/KITTI, MASt3R/BlendedMVS) have $R^2\le0.05$,
that is, no detectable distance dependence at all rather than a contradicting one. This extends the within-scene
test beyond VGGT to all three backbones.

An additional test does not establish whether the \emph{magnitude} of the recovered contamination predicts how
much alignment helps. Regressing the AUSE reduction on
$(\sigma_s^2+2\sigma_r^2)\,\mathbb{E}\|x\|^2/\mathbb{E}[U_{\mathrm{aligned}}]$ over the $16$ conditions with a
positive recovered coefficient gives Spearman $+0.03$, or $-0.20$ when the reduction is first divided by the
built-in AUSE to account for the headroom the built-in confidence leaves. Neither is informative at this sample
size: the bootstrap $95\%$ interval on the correlation is $[-0.49,+0.59]$ and a permutation test gives $p=0.91$, so
the design neither supports nor excludes a magnitude relation. The result is reported because the analysis is specified
before execution and shares its fitted coefficients with the result above, not because it settles the
question. Eq.~\eqref{eq:decomp} predicts the \emph{form} of the contamination; how much a ranking metric improves
once that contamination is removed depends further on how much of the remaining error the perturbation family
exposes, and on how much headroom the baseline leaves.

The effect is expected to scale with
how far a backbone's output frame actually moves between passes, and the pose modality supports that reading: the
analogous $\|c\|^2$ signature is strong on VGGT and statistically absent on the permutation-equivariant \pithree{}
(Table~\ref{tab:poses}). The confound is therefore backbone-dependent in magnitude, not universal in strength.

The gauge/covariance coupling is classical in bundle adjustment~\citep{triggs2000}; here it identifies the dominant
nuisance in perturbation-ensemble uncertainty for frozen point maps.

\subsection{Second Modality: SE(3) Camera Poses}
\label{app:poses}

Camera poses exhibit the same gauge contamination, although alignment does not make pose instability a uniformly
strong error ranker. Equation~\eqref{eq:decomp} applies to outputs with an unobserved global symmetry, not only to
point maps. Feed-forward reconstruction places points and cameras in a shared frame that floats in
$\mathrm{Sim}(3)$ across passes. This second structured output is tested by reusing the point maps' per-run similarity
fit to gauge-align the per-view camera poses.

\paragraph{Protocol.} Each view-set uses the main estimator's $N{=}5$ photometric passes, from which the
point map and per-view camera pose are collected. For run $n$, the same code as Section~\ref{sec:method} fits one global robust
similarity $(s_n,R_n,t_n)$ from its point map to the reference, then applies that transform to its cameras. Camera
centres follow $c\mapsto s_nR_nc+t_n$, and camera-to-world rotations follow $R_c\mapsto R_nR_c$. \emph{Naive} pose
uncertainty is the cross-run spread before this correction; \emph{gauge-aligned} uncertainty is the spread after it.
Centre uncertainty is defined as the trace spread of camera centres and rotation uncertainty as the RMS geodesic
angle about their chordal mean. Pose error is the reference pose's centre distance and geodesic rotation angle after
one gauge-free similarity alignment to ground-truth poses. AUSE and Spearman pool all cameras ($120$ to $150$ per
condition). The pose convention is determined once against ground truth: camera-to-world for \pithree{} and
world-to-camera for VGGT.

\paragraph{Result.} The $\|x_p\|^2$ signature of Eq.~\eqref{eq:decomp} transfers to camera centres, although the
error-ranking gains it produces are mixed. Table~\ref{tab:poses} shows that naive centre uncertainty grows with camera-centre magnitude $\|c\|^2$, strongly on
VGGT (Spearman $0.61$ to $0.70$, $p<10^{-12}$ on every dataset) and on \pithree{}/7-Scenes, but only weakly on the
permutation-equivariant \pithree{} elsewhere. This pattern tracks how much each backbone's frame moves between
passes. Gauge removal lowers centre pose-error AUSE in all six conditions. However, the reported AUSE risk is
outlier-sensitive; under the outlier-robust rank correlation, centre improvement holds in only three of six (both
DTU conditions and \pithree{}/7-Scenes). On VGGT/7-Scenes and VGGT/ETH3D, centre uncertainty remains a weak error
ranker with or without alignment. Rotation AUSE improves in four of six, including all three VGGT conditions. These
results are interpreted as evidence that Eq.~\eqref{eq:decomp}'s contamination is present and removable in a second
structured output, most clearly for the non-equivariant backbone whose frame moves. No pose-uncertainty method is
proposed.

\begin{table*}[t]
\caption{\textbf{The gauge confound transfers to SE(3) camera poses, but ranking gains are mixed.} Naive and
gauge-aligned pose uncertainty rank pose error over all cameras pooled per condition (AUSE lower is better; Spearman
higher is better). Bold marks the better AUSE. Centre alignment lowers AUSE in all six conditions, whereas
outlier-robust centre Spearman improves in three (both DTU and \pithree{}/7-Scenes); VGGT/7-Scenes and VGGT/ETH3D
remain weak negative rankers. The final column measures Eq.~\eqref{eq:decomp}'s $\|c\|^2$ signature through the rank
correlation between naive centre uncertainty and squared camera-centre magnitude, which is strong where the frame
moves.}
\label{tab:poses}
\begin{center}\begin{tabular}{ll cc cc cc c}
\toprule
& & \multicolumn{2}{c}{\textbf{Centre AUSE}} & \multicolumn{2}{c}{\textbf{Centre Spearman}} & \multicolumn{2}{c}{\textbf{Rot.\ AUSE}} & $\pmb{\rho(\|c\|^2)}$ \\
\cmidrule(lr){3-4}\cmidrule(lr){5-6}\cmidrule(lr){7-8}
\textbf{Backbone} & \textbf{Dataset} & naive & align & naive & align & naive & align & naive \\
\midrule
\tblposes
\bottomrule
\end{tabular}
\end{center}
{$^{\ast}\,p<10^{-4}$. Centre Spearman is negative on VGGT/7-Scenes and VGGT/ETH3D for both naive and
aligned, indicating pose-centre instability is a weak proxy for pose error there regardless of the gauge.}
\end{table*}

\section{Estimator Components and Alignment}
\label{app:ablations}

Alignment removes the dominant nuisance in the training-free ranking, while held-out fusion supplies a smaller,
more consistent gain and calibration affects magnitude rather than rank. Algorithm~\ref{alg:revar3r} gives the
complete procedure of Section~\ref{sec:method}.

\begin{algorithm}[htb]
\SetAlgoLined
\KwIn{frozen model $f$; $V$ input views; passes per family $N$; perturbation families $\mathcal{F}$, each
$F\in\mathcal{F}$ a distribution over label-preserving input transforms $P_n^{F}$}
\KwOut{per-point uncertainty $u[p]\ge 0$}
$X^{(0)}\leftarrow f(\text{views})$ \tcp*{reference prediction (gauge reference frame)}
$\tilde{X}^{(0)}\leftarrow X^{(0)}$ \tcp*{reference needs no alignment}
\ForEach{family $F\in\mathcal{F}$}{
 \For{$n=1$ \KwTo $N-1$}{
 $X^{(n)}\leftarrow f\big(P_n^{F}(\text{views})\big)$ \tcp*{re-run on perturbed input}
 $\hat{T}_n\leftarrow \textsc{RobustSim3}\big(X^{(n)},X^{(0)}\big)$;\quad
 $\tilde{X}^{(n)}\leftarrow \hat{T}_n(X^{(n)})$ \tcp*{align before variance}
 }
 $u_F[p]\leftarrow \operatorname{tr}\,\mathrm{Cov}_n\big[\tilde{X}^{(n)}[p]\big]$ \tcp*{aggregate over runs}
}
$u[p]\leftarrow \textsc{RankFuse}\big(\{u_F[p]\}_{F\in\mathcal{F}}\cup\{\text{built-in conf.}\}\big)$
\tcp*{optional, held-out weights}
\caption{\methodname{}: gauge-aware perturbation uncertainty (single view $v$ shown; run over all $V$). Here
$\hat{T}_n\in\Rthree{}$ is the robust similarity fit of Section~\ref{sec:robust}, $\tilde{X}^{(n)}$ the aligned
prediction, $\mathrm{Cov}_n$ and $\operatorname{tr}$ the per-point sample covariance across the $N$ runs and its
trace, and $\textsc{RankFuse}$ the held-out inverse-AUSE$^2$ rank fusion of Section~\ref{sec:aggfuse}.}
\label{alg:revar3r}
\end{algorithm}

\begin{table*}[t]
\caption{\textbf{The three levels of \methodname{}.} The training-free core produces the AUSE/Spearman ranking.
Fusion and calibration use a small held-out labelled split; neither changes a single family's ranking, and
calibration alone changes only magnitude. This separation distinguishes ranking from calibration in the Trust3R
comparison (Section~\ref{sec:trust3r}).}
\label{tab:levels}
\begin{center}\begin{tabular}{lll}
\toprule
\textbf{Level} & \textbf{What it adds} & \textbf{Labels} \\
\midrule
Core & aligned single-family variance (a ranking) & none \\
$+$ fusion & rank-space fusion of families with the built-in confidence & held-out split \\
$+$ calibration & isotonic map from score to error magnitude (ECE) & held-out split \\
\bottomrule
\end{tabular}
\end{center}
\end{table*}

\begin{figure*}[t]
\centering
\includegraphics[width=\textwidth]{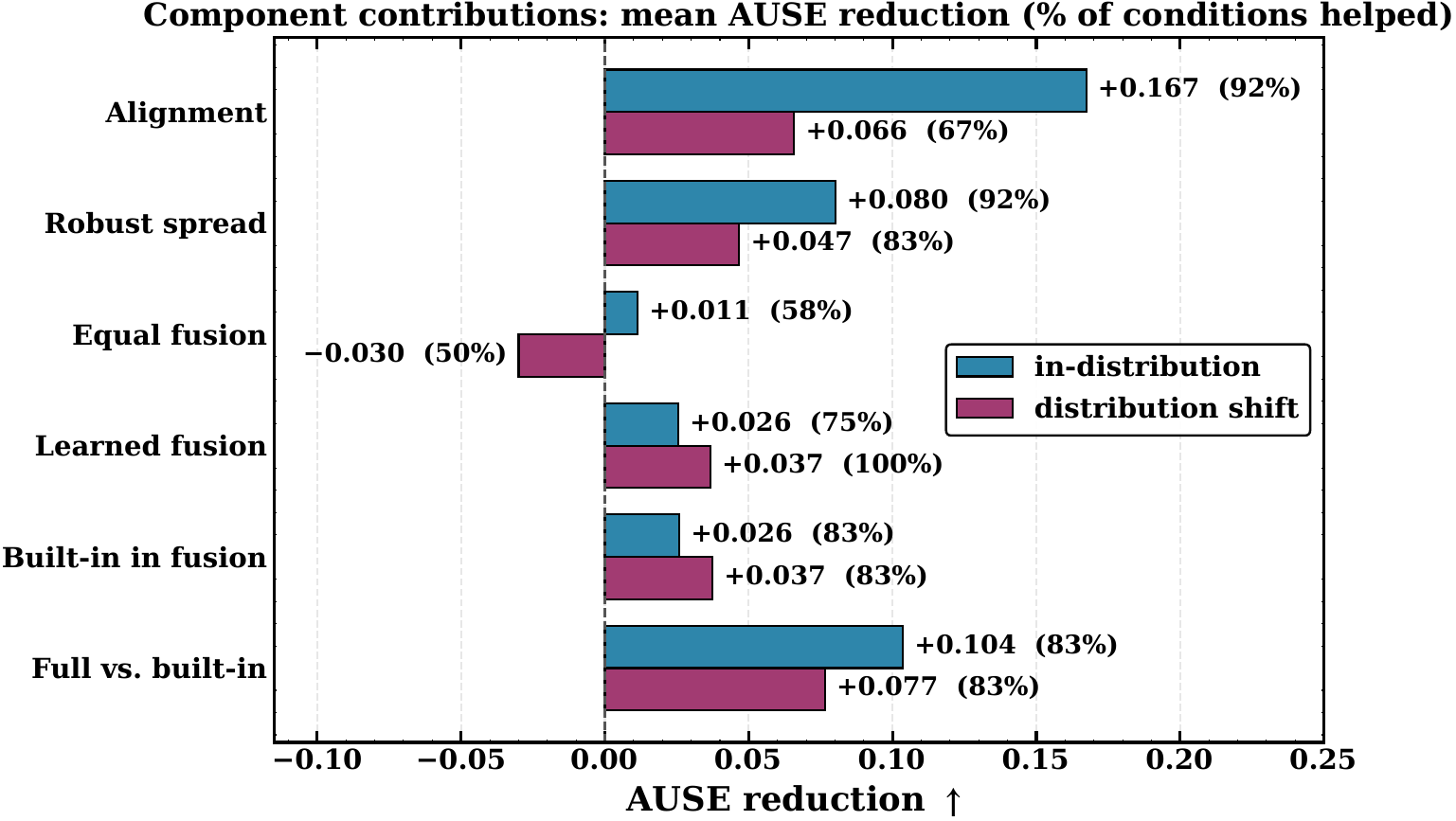}
\caption{\textbf{Component decomposition.} Mean AUSE reduction (positive $=$ improvement) and, in parentheses, the share of
conditions each component helps, split into in-distribution and shifted conditions. Alignment supplies the largest
in-distribution gain, held-out learned fusion is the most consistent component, and equal fusion exhibits occasional degradation under shift.
The fifth group isolates admitting built-in confidence as a fusion input: it is of the same order as the fusion step
itself, which is why the label-free core ($11/18$) is reported alongside the fused $15/18$.}
\label{fig:decomp}
\end{figure*}

The component ladder shows that alignment dominates in distribution, held-out learned fusion is the most consistent
step overall, and naive equal fusion is unreliable. A ten-step progression is evaluated across $3$ models $\times$
$6$ datasets: built-in confidence; individual perturbation families without alignment; alignment; variance versus
robust spread; equal fusion; held-out learned fusion; and the full estimator. Figure~\ref{fig:decomp} gives each
step's mean AUSE change and help rate for in-distribution (ID) and distribution-shift (OOD) conditions. After
alignment, photometric and input-noise perturbations differ only marginally in best-configuration AUSE on average.
Their choice matters more under shift, where input noise is marginally stronger.

Robust Umeyama alignment reduces AUSE on every backbone when averaged over the six datasets ($15/18$ conditions
individually; see Figure~\ref{fig:decomp}): $0.165$, $0.131$, and $0.105$ on VGGT, MASt3R, and \pithree{},
respectively. It is at least as good as plain Umeyama on every
backbone. Figure~\ref{fig:decomp} shows that alignment dominates in distribution, held-out learned fusion alone helps
in every shifted condition, and naive equal fusion is unreliable.

The training-free core, not fusion with built-in confidence, drives the gain. Table~\ref{tab:coredecomp} separates
aligned single-family variance, with neither fusion nor built-in confidence, from the fused variant. The core lowers
AUSE below built-in confidence in $8/12$ conditions on the four original datasets and $11/18$ across all six
(Table~\ref{tab:newds}), compared with only $5/18$ for the same perturbation \emph{without} alignment. The optional
held-out fusion recovers additional cells, especially where one family is weak, but does not create the main result.

\begin{table}[tb]
\caption{\textbf{Training-free core vs.\ built-in.} Decomposition of the gain (AUSE~\down{}) across the four original
datasets (twelve conditions): built-in confidence; \emph{unaligned}
single-family perturbation variance; the \emph{training-free core} \methodname{} estimator (robust-aligned
single-family variance, no fusion, no built-in confidence); and the full held-out fusion. The training-free core
beats built-in in $8/12$ here and $11/18$ with the competitor benchmarks (Table~\ref{tab:newds}), versus $5/18$ for
unaligned perturbation. The gain comes from alignment, not fusion with the baseline.}
\label{tab:coredecomp}
\begin{center}\scriptsize\setlength{\tabcolsep}{3pt}\begin{tabular}{ll cccc}
\toprule
\textbf{Model} & \textbf{Dataset} & \textbf{built-in} & \textbf{unaligned} & \textbf{core (train-free)} & \textbf{full (fused)} \\
\midrule
\tblcoredecomp
\bottomrule
\end{tabular}
\end{center}
\end{table}

\paragraph{Aggregation statistic.} Trace variance is the strongest aggregator. Table~\ref{tab:aggabl} compares the
three choices from Section~\ref{sec:aggfuse}, trace covariance, median-absolute-deviation spread, and largest
covariance eigenvalue, on aligned photometric spread across the same $3\times6$ grid. The trace column reproduces
the training-free core in Table~\ref{tab:coredecomp}. Median spread is \emph{clearly} weaker (mean AUSE $+0.068$,
weaker in $16/18$ conditions), whereas the largest eigenvalue is only \emph{marginally} weaker (mean $+0.006$,
weaker in $15/18$ but close to a tie).

\begin{table}[htb]
\caption{\textbf{Aggregation statistic (full grid).} Mean AUSE over $3$ models $\times$ $6$ datasets for three
aligned-photometric aggregators; lower is better. Trace variance is strongest, median spread is clearly weaker, and
the largest eigenvalue is marginally weaker. The trace column reproduces the rung-7 core values of
Table~\ref{tab:coredecomp} on the four original datasets; TUM and KITTI are recomputed here over ten view-sets
rather than the eight used in Table~\ref{tab:newds}.}
\label{tab:aggabl}
\begin{center}\begin{tabular}{lccc}
\toprule
\textbf{Aggregator} & \textbf{ID} & \textbf{OOD} & \textbf{overall} \\
\midrule
\tblaggabl
\bottomrule
\end{tabular}
\end{center}
\end{table}

\subsection{Alignment Ablations: Reference, Group, and Scope}
\label{app:alignabl}
The alignment ablations establish three facts: the similarity group is necessary, the reference choice is
immaterial, and per-view scope outperforms scene-level scope. Table~\ref{tab:alignabl} varies only alignment while
sharing forward passes, then averages the photometric core over the eighteen model--dataset conditions.

\textbf{Group.} AUSE falls monotonically as the group expands. The two largest steps add rotation
(translation$\to$rigid) and scale (rigid$\to$similarity), removing precisely the
$(\sigma_s^2+2\sigma_r^2)\|x\|^2$ term that Eq.~\eqref{eq:decomp} identifies as the dominant distance-scaling
nuisance; translation, a position-independent offset in the theory, helps least.

\textbf{Reference.} Fixed run~$0$,
ensemble medoid, and random-run anchors are indistinguishable, so the signal belongs to the ensemble rather than a
privileged reference.

\textbf{Scope.} Independent per-view alignment beats a single scene-level transform, which
cannot absorb per-view gauge variation. The scene-level transform removes the predicted global gauge; per-view
alignment additionally stabilizes residual view-dependent frame variation outside the proposition's single-transform
model. Each view's error is evaluated after its own similarity alignment to ground truth, so these offsets are absent
from the target and removing them does not score away error. The estimator is interpreted as global gauge
removal followed by evaluation-consistent per-view frame stabilization, not as a claim that the entire improvement
follows from Proposition~\ref{prop:decomp}.

\begin{table}[tb]
\caption{\textbf{Alignment ablations: group, reference, and scope.} Photometric-core means over $18$ conditions
(AUSE~\down{}, Spearman~\up{}). Similarity alignment performs best, reference choices are indistinguishable, and
per-view scope beats scene-level scope. ``helps'' counts conditions beating built-in AUSE;
similarity/run-0/per-view is the default core and appears identically in all three blocks. This ablation uses the
first $8$ view-sets in every condition, whereas Table~\ref{tab:coredecomp} uses $10$ for the four original datasets
(TUM and KITTI have eight in both). With the larger sample, unaligned variance beats built-in confidence in $5/18$
conditions rather than $3/18$: \pithree{} on ETH3D and VGGT on BlendedMVS change sign.}
\label{tab:alignabl}
\begin{center}\scriptsize\setlength{\tabcolsep}{3pt}\begin{tabular}{ll cccc c}
\toprule
 & & \multicolumn{3}{c}{\textbf{AUSE}} & & \\
\cmidrule(lr){3-5}
\textbf{axis} & \textbf{variant} & \textbf{all} & \textbf{ID} & \textbf{OOD} & \textbf{Spearman} & \textbf{helps} \\
\midrule
\tblalignabl
\bottomrule
\end{tabular}
\end{center}
\end{table}

\begin{figure*}[t]
\centering
\includegraphics[width=\textwidth]{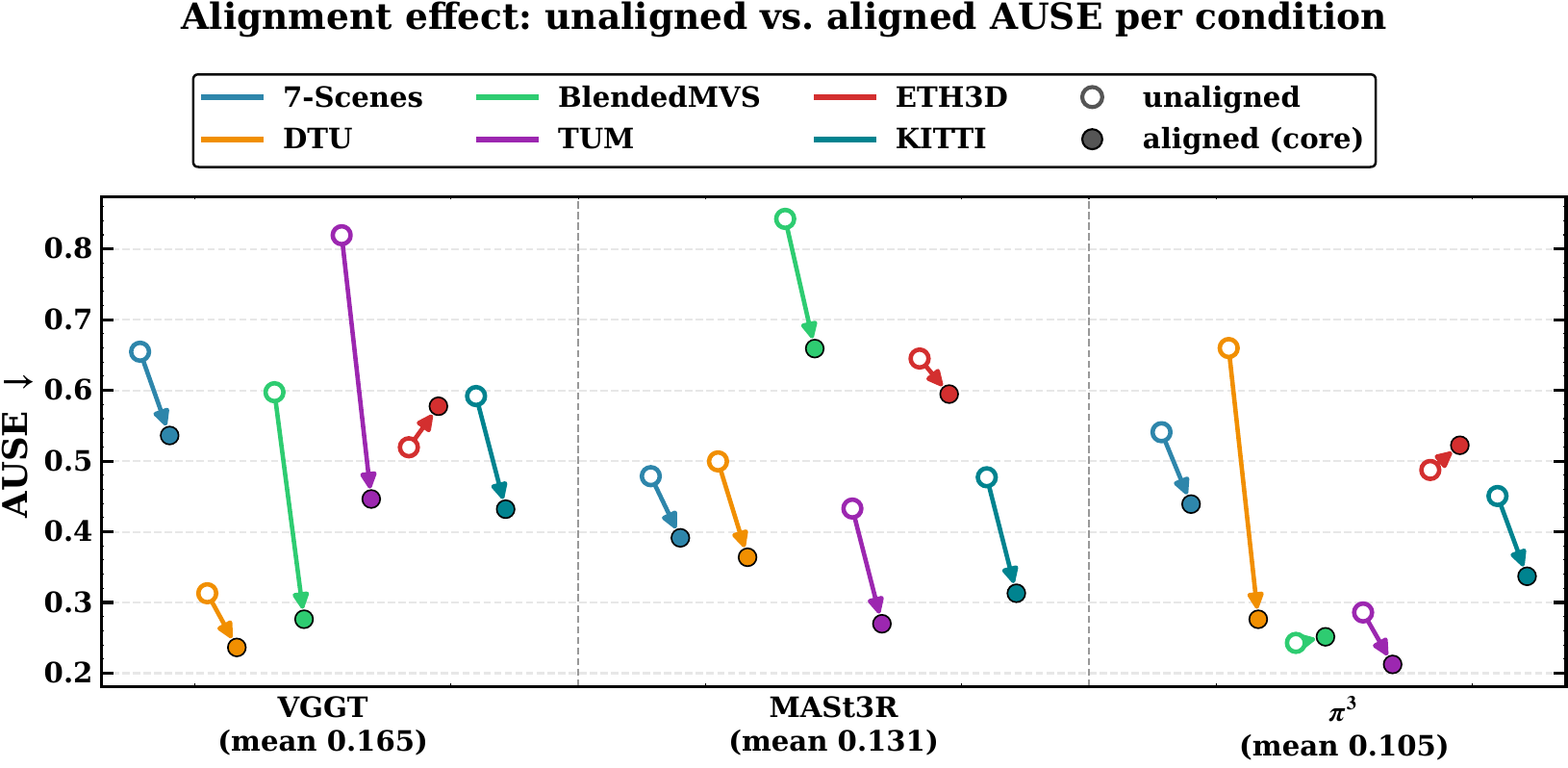}
\caption{\textbf{Alignment removes the dominant nuisance.} For each backbone (group) and dataset (colour), AUSE~\down{}
of photometric perturbation variance without alignment (hollow circle) and of the same runs after gauge alignment
(filled circle, the \methodname{} core); each arrow runs from unaligned to aligned, so an arrow sloping down is a
reduction. The pair of one condition is offset horizontally so the arrow stays readable where the effect is small. The labels under the groups give the mean reduction over the six datasets (VGGT $0.165$, MASt3R
$0.131$, \pithree{} $0.105$). Alignment lowers AUSE in $15$ of the $18$ conditions; the exceptions
are ETH3D for VGGT and \pithree{}, and BlendedMVS for \pithree{}, the last by $0.009$. Table~\ref{tab:coredecomp}
lists the values for the four original datasets.}
\label{fig:align}
\end{figure*}

\paragraph{Aligned sampling baselines.} Alignment, not the perturbation recipe, carries the result. The
input-noise (Monte-Carlo) ensemble is rerun through the identical registration, aggregation and scoring path used for
photometric jitter, at the same pass budget, and both are compared against built-in confidence.
Table~\ref{tab:alignedmc} reports the outcome: aligned input-noise lowers AUSE below built-in confidence in
\textbf{all nine} conditions, whereas the same ensemble without alignment does so in five, and aligned input-noise
is closely matched to aligned photometric jitter in this nine-condition comparison (lower in five conditions, higher in four).
These results indicate that the transferable content of this paper is the alignment step, and that the choice of
label-preserving perturbation is a second-order implementation detail.

\begin{table*}[t]
\caption{\textbf{Aligned sampling baseline at matched budget.} AUSE~\down{} for the input-noise ensemble with and
without gauge alignment, beside aligned photometric jitter, all scored identically. Bold marks the best of the three
uncertainty estimators per row. Same pass budget, registration and scoring throughout, so the columns differ only
in whether the ensemble is aligned and in which perturbation generates it.}
\label{tab:alignedmc}
\begin{center}\begin{tabular}{ll c ccc}
\toprule
 & & & \multicolumn{2}{c}{\textbf{input-noise}} & \textbf{photometric} \\
\cmidrule(lr){4-5}\cmidrule(lr){6-6}
\textbf{Model} & \textbf{Dataset} & \textbf{built-in} & unaligned & aligned & aligned \\
\midrule
\tblalignedmc
\bottomrule
\end{tabular}
\end{center}
\end{table*}

\section{Evaluation and Comparisons}

\subsection{Per-Condition Results}
\label{app:perscene}

Across the twelve conditions, \methodname{} improves Spearman in eight, ties in one
(\pithree{}/ETH3D, $0.19\to0.19$), and regresses in three (\pithree{}/BlendedMVS $0.59\to0.48$,
MASt3R/ETH3D $0.27\to0.22$, MASt3R/BlendedMVS $0.50\to0.35$), all cells where the built-in ranking is
already comparatively strong. Table~\ref{tab:perscene} supplies the exact AUSE cells behind the main heatmap
(Figure~\ref{fig:mainheat}) and the corresponding Spearman values for built-in confidence and full \methodname{}.
The Trust3R comparison is reported separately in Appendix~\ref{app:reconcile}, where the scene structure of these
benchmarks is discussed and effect sizes are given in place of significance tests.

\begin{table*}[t]
\caption{\textbf{Per-condition results at native resolution.} AUSE (\down{}) and Spearman (\up{}) compare built-in
confidence with full \methodname{}. Improvements concentrate where built-in ranking is weak; regressions occur
where it is already comparatively strong.}
\label{tab:perscene}
\begin{center}\begin{tabular}{ll cc cc}
\toprule
 & & \multicolumn{2}{c}{\textbf{AUSE~\down{}}} & \multicolumn{2}{c}{\textbf{Spearman~\up{}}} \\
\cmidrule(lr){3-4}\cmidrule(lr){5-6}
\textbf{Model} & \textbf{Dataset} & \textbf{built-in} & \textbf{\methodname{}} & \textbf{built-in} & \textbf{\methodname{}} \\
\midrule
\tblperscene
\bottomrule
\end{tabular}
\end{center}
\end{table*}

\subsection{Competitor-Benchmark Results (TUM, KITTI)}
\label{app:newds}
The full estimator beats built-in AUSE in all six competitor-benchmark conditions, while the training-free core is
more effective on far-out-of-domain KITTI than indoor TUM. Table~\ref{tab:newds} decomposes both Trust3R benchmarks
at each backbone's native resolution, matching the main heatmap (Figure~\ref{fig:mainheat}). The core wins $1/3$ on
TUM, where built-in confidence is strong, and $2/3$ on KITTI. On KITTI it also raises Spearman to $+0.67$ for the
core and $+0.77$ for the full estimator, compared with $+0.30$--$+0.53$ for built-in confidence. Table~\ref{tab:trust3r}
gives the head-to-head comparison with the trained Trust3R head.

\begin{table*}[t]
\caption{\textbf{Competitor benchmarks (TUM, KITTI).} AUSE~\down{}~/~Spearman~\up{} for built-in confidence,
\emph{unaligned} single-family perturbation, the \emph{training-free core} (aligned single-family variance), and the
full fused estimator. Full \methodname{} beats built-in AUSE in $6/6$ conditions; the core wins in $3/6$
(\pithree{} on both benchmarks and MASt3R/KITTI).}
\label{tab:newds}
\begin{center}\begin{tabular}{ll cccc}
\toprule
\textbf{Dataset} & \textbf{Model} & \textbf{built-in} & \textbf{unaligned} & \textbf{core (train-free)} & \textbf{full (fused)} \\
\midrule
\tblnewds
\bottomrule
\end{tabular}
\end{center}
\end{table*}

\subsection{Comparison with a Trained Uncertainty Head}
\label{app:reconcile}

This section gives the full comparison against Trust3R~\citep{trust3r}, the closest learned alternative, and states
its limits. The comparison is reported as a trade-off, and no claim in the paper rests on it.

\paragraph{Protocol.} Trust3R's released checkpoint is a MASt3R-based implementation at $224$~px, so the comparison
is available on that backbone only. Both methods are scored inside the present evaluation harness on identical view-sets, identical
ground truth, the same gauge-free alignment and the same AUSE definition, which is what makes them comparable to
each other. Trust3R uses its paper-default \emph{epistemic} readout for ranking and the full posterior-predictive
variance for calibration. All available view-sets per dataset are used; Trust3R also reports ScanNet++, which
is unavailable for the present study.

\paragraph{Scenes, not view-sets, are the unit of inference.} View-sets drawn from one scene are correlated, so a
paired test over view-sets is anticonservative. Blocking by scene, the smallest attainable two-sided $p$ is
$2/2^{G}$ for $G$ scenes, so the number of \emph{independent scenes} sets a hard floor on the significance attainable by any test.
One view-set per scene is therefore sampled wherever a dataset supplies enough scenes, and an exact
cluster sign-flip permutation tests per-scene mean differences.

On DTU this is decisive. Sampling one view-set from each of $30$ scans
instead of ten view-sets from three scans, \methodname{} attains a lower per-view-set AUSE than the trained head on
\textbf{every one of the $30$ scans}: median $\mathbf{0.228}$ against $0.459$, per-scene mean difference
$\mathbf{-0.205}$, and the closest scan still favours \methodname{} by $0.027$. The scene-level sign-flip test
gives $p=1.9\times10^{-9}$, the smallest value attainable with $30$ scans, which survives Holm correction over the
three scene-blocked tests. Pooled AUSE agrees on the same sample ($0.403$ against $0.637$), so the result does not
depend on the aggregation. Seed spread of per-scan AUSE is $0.025$ against an effect of $0.205$.

ETH3D supplies nine scenes and 7-Scenes seven rooms (floors $3.9\times10^{-3}$ and $1.6\times10^{-2}$), and
neither reaches significance: the per-scene mean differences are $-0.034$ and $-0.027$, favouring \methodname{} in
$5/9$ scenes and $4/7$ rooms, with sign-flip $p=0.23$ and $p=0.64$. KITTI and TUM cannot be
tested at all, supplying a single drive and two sequences. Only one significance claim is therefore made, on the dataset
where the effect is large and the scene count supports it; the remaining results are reported as effect sizes.

\paragraph{Result.} Table~\ref{tab:trust3r} reports both aggregations. They disagree in several
conditions, and the disagreement is informative. Pooled AUSE concatenates points across view-sets and is
therefore dominated by the largest-error scenes, whereas the per-view-set median is robust to scene
heterogeneity and is the quantity examined by the paired analysis. Under the per-set median \methodname{} is
favoured on DTU, ETH3D and KITTI; under pooled AUSE the trained head leads on 7-Scenes, TUM and ETH3D, while
\methodname{} leads on DTU and KITTI. Trust3R is the better magnitude calibrator throughout. The consistent
reading across both aggregations is that the trained head is stronger inside its training domain and better
calibrated, while the training-free estimator is favoured on the most dissimilar data and transfers to any
backbone without adaptation.

\begin{table*}[t]
\caption{\textbf{\methodname{} vs.\ a trained evidential head (Trust3R).}
Frozen MASt3R at $224$~px on identical point sets and all available view-sets. AUSE~\down{},
Spearman~\up{}; Trust3R uses its paper-default \emph{epistemic} readout and \methodname{}
entries are means over five seeds. ``scenes'' counts the independent scenes available for inference.}
\label{tab:trust3r}
\begin{center}
\begin{tabular}{ll ccccc}
\toprule
 & & \textbf{7-Scenes} & \textbf{TUM} & \textbf{ETH3D} & \textbf{DTU} & \textbf{KITTI} \\
\midrule
\multicolumn{2}{l}{sets} & 23 & 12 & 24 & 30 & 12 \\
\multicolumn{2}{l}{scenes} & 7 & 2 & 9 & 30 & 1 \\
\midrule
pooled AUSE~\down{} & built-in & 0.535 & 0.232 & 0.585 & 0.557 & 0.359 \\
 & \methodname{} & 0.293 & 0.188 & 0.667 & \textbf{0.403} & \textbf{0.216} \\
 & Trust3R & \textbf{0.268} & \textbf{0.119} & \textbf{0.477} & 0.637 & 0.319 \\
\midrule
per-set median AUSE~\down{} & \methodname{} & 0.272 & 0.148 & \textbf{0.471} & \textbf{0.228} & \textbf{0.209} \\
 & Trust3R & \textbf{0.271} & \textbf{0.125} & 0.514 & 0.459 & 0.242 \\
\midrule
Spearman~\up{} & \methodname{} & $\mathbf{+0.43}$ & $+0.59$ & $+0.13$ & $\mathbf{+0.43}$ & $\mathbf{+0.69}$ \\
 & Trust3R & $+0.36$ & $\mathbf{+0.69}$ & $\mathbf{+0.23}$ & $-0.13$ & $+0.44$ \\
\bottomrule
\end{tabular}
\end{center}
\end{table*}

\paragraph{Sensitivity to the view-set count.} An earlier protocol used ten view-sets per dataset (eight for TUM and
KITTI). On TUM, ETH3D and KITTI the present sample extends that list. On the shared view-sets the present run
reproduces Trust3R's per-set AUSE exactly and \methodname{}'s mean AUSE to within $0.01$, although individual
view-sets move by up to $0.043$, so the two protocols agree on average where they overlap. The added
view-sets are not uniformly harder: they are easier on ETH3D and TUM and harder on KITTI. On 7-Scenes and DTU the
present samples are instead drawn across seven rooms and thirty scans, so they replace rather than extend the
earlier view-sets. The pooled numbers nonetheless move substantially, which is the aggregation effect described
above rather than a difference in difficulty.

\paragraph{Relation to Trust3R's published table.} The reported numbers are not directly comparable to the values published
with Trust3R, and no reproduction is claimed. The metric differs: they report the unnormalized area between
the risk-coverage curve and its oracle, in scene units, whereas the present evaluation reports AUSE after per-scene normalization. The
evaluation also differs in sample ($5000$ image pairs drawn from their benchmark splits, against the present $V{=}5$ view-sets),
in grouping, and in the masking and alignment path, since every method here is scored through the present evaluation harness. As a
partial check, recomputing their metric definition on the present point sets and dividing by mean error to remove the
scene-scale factor gives $0.259$ on KITTI against their published $0.266$, while TUM gives $0.127$ against their
published $0.204$; rank correlation, which is normalization-invariant, agrees on KITTI ($+0.44$ against $+0.46$)
and differs on ETH3D, where the present wide-baseline view-sets leave every method near zero. Table~\ref{tab:aurc} reports
their metric on the present point sets. The released code also computes the epistemic readout as
$1/(\kappa(\nu-2))$, which is implemented by the evaluation wrapper, rather than the $\kappa(\nu-4)$ given in its
documentation.

\begin{table*}[t]
\caption{\textbf{Trust3R's metric definition, recomputed on the present point sets.} Risk-coverage AURC~\down{} and
\emph{unnormalized} AUSE~\down{}, both in scene units, so methods are comparable within a row but values are not
comparable across datasets. Ten view-sets per dataset (eight for TUM and KITTI), seed $0$, Trust3R \emph{total}
readout, and an exact sort-based curve in place of their log-binned estimator; it is a cross-check on the metric
definition, not a reproduction of their protocol.}
\label{tab:aurc}
\begin{center}\begin{tabular}{l cccc cccc}
\toprule
 & \multicolumn{4}{c}{\textbf{AURC~\down{}}} & \multicolumn{4}{c}{\textbf{AUSE~\down{} (unnormalized)}} \\
\cmidrule(lr){2-5}\cmidrule(lr){6-9}
 & built-in & \methodname{} & input & Trust3R & built-in & \methodname{} & input & Trust3R \\
\textbf{Dataset} & & & noise & & & & noise & \\
\midrule
\tblaurc
\bottomrule
\end{tabular}
\end{center}
\end{table*}

\subsection{Frozen-Feature Probe}
\label{app:probe}
The frozen-feature probe has worse AUSE than \methodname{} on all three tested datasets, but better Spearman on DTU,
so the two rankings are not uniformly ordered. Table~\ref{tab:probe} compares \methodname{} with a learned VGGT probe
trained to predict per-point error from frozen features. This is a separate VGGT run that scores the
view-re-ordering family rather than the photometric core on the main protocol, so its built-in-confidence column
differs from Table~\ref{tab:perscene} (most visibly on DTU, $0.402$ vs.\ $0.248$); read the two tables
independently rather than cell by cell. On DTU, the probe reaches Spearman $0.56$ versus $0.46$ for
\methodname{}. Overall, the result remains consistent with cross-run input sensitivity exposing reliability
information that a single deterministic pass does not fully recover.

\begin{table}[tb]
\caption{\textbf{Frozen-feature probe on VGGT.} AUSE~\down{}~/~Spearman~\up{} for built-in confidence, a learned
probe, and \methodname{}. \methodname{} has lower AUSE throughout, while the probe has higher Spearman on DTU.}
\label{tab:probe}
\begin{center}\scriptsize\setlength{\tabcolsep}{3pt}\begin{tabular}{lccc}
\toprule
\textbf{Signal} & \textbf{7-Scenes} & \textbf{DTU} & \textbf{ETH3D} \\
\midrule
\tblprobe
\bottomrule
\end{tabular}
\end{center}
\end{table}

\subsection{Downstream Utility}
\begin{figure*}[t]
\centering
\includegraphics[width=\textwidth]{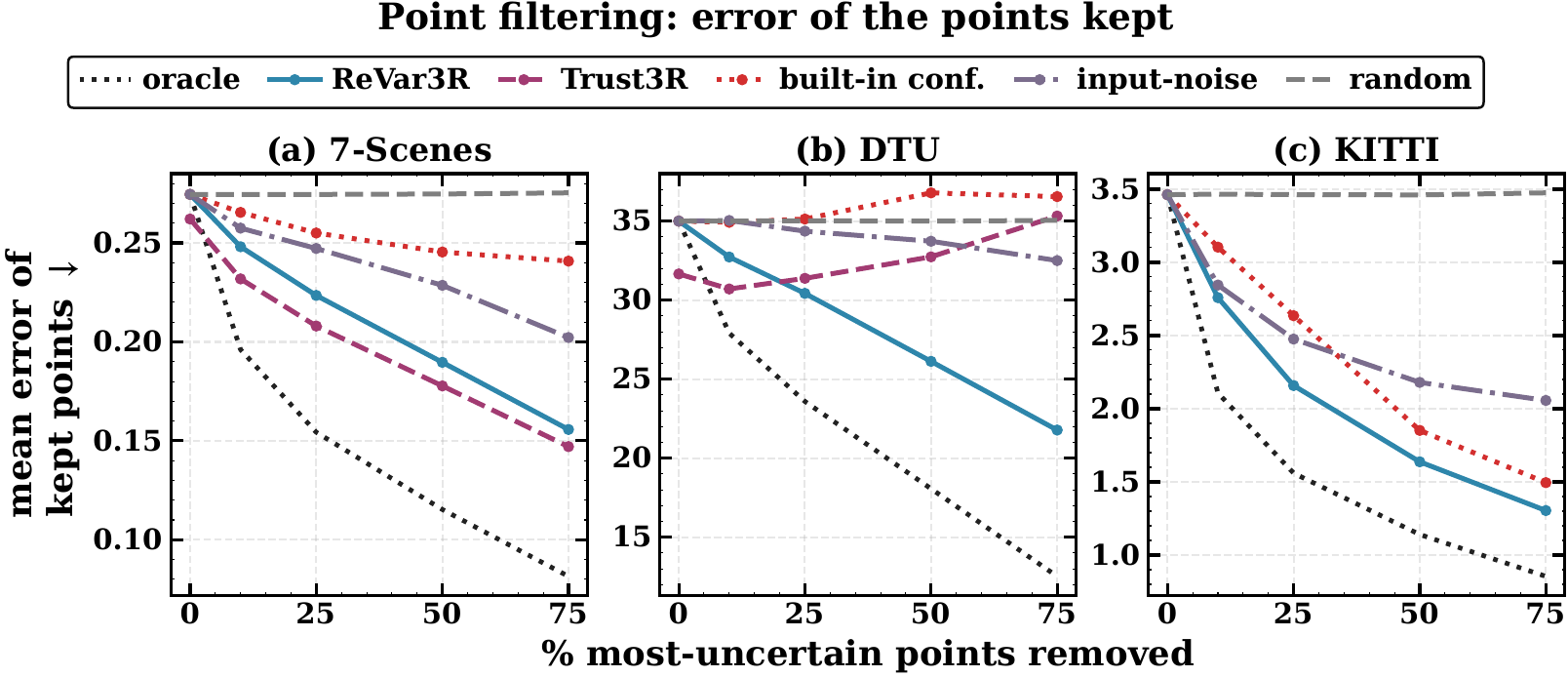}
\caption{\textbf{Better ranking yields better geometry.} Mean error of kept points~\down{} vs.\ removal fraction.
Dropping the points \methodname{} ranks most-uncertain retains lower-error geometry than built-in confidence, a
naive input-noise ensemble, or random removal, closing $30$--$57\%$ of the built-in-to-oracle gap at $50\%$
removal. All panels use frozen MASt3R: (a) and (b) the $224$~px protocol with ten view-sets, (c) the
native-resolution KITTI run with eight, where the $224$~px Trust3R checkpoint does not apply. Trust3R filters its
own refined point map and starts from a different zero-removal error (DTU $31.7$ vs.\ $35.0$), so read its shape,
not its offset.}
\label{fig:downstream}
\end{figure*}

Figure~\ref{fig:downstream} shows the full point-filtering curves summarized in Section~\ref{sec:downstream}.

\subsection{Negative Results: Systematic Bias, Calibration Transfer, NVS}
\label{app:negatives}

Three negative results bound the estimator: variance misses stable systematic bias, magnitude calibration does not
transfer across domains, and uncertainty-guided initialization does not improve novel-view synthesis.

\paragraph{Stable systematic bias.} Variance does not detect errors that remain stable across perturbations. In a
controlled mixture of unstable points, whose predictions disagree, and stably biased points, whose predictions
agree on a wrong value, the variance ranking inverts as biased fraction $f$ grows: Spearman starts at $+0.39$ for
$f{=}0$ and falls to $-0.67$ at $f{=}0.6$. On real data, top-decile-error detection AUROC is $0.75$ on 7-Scenes and
$0.80$ on DTU, but near chance under shift ($0.53$ on ETH3D). A trained head does not resolve the limitation
(e.g.\ AUROC $0.49$ on DTU).

\begin{figure}[htb]
\centering
\includegraphics[width=0.8\linewidth]{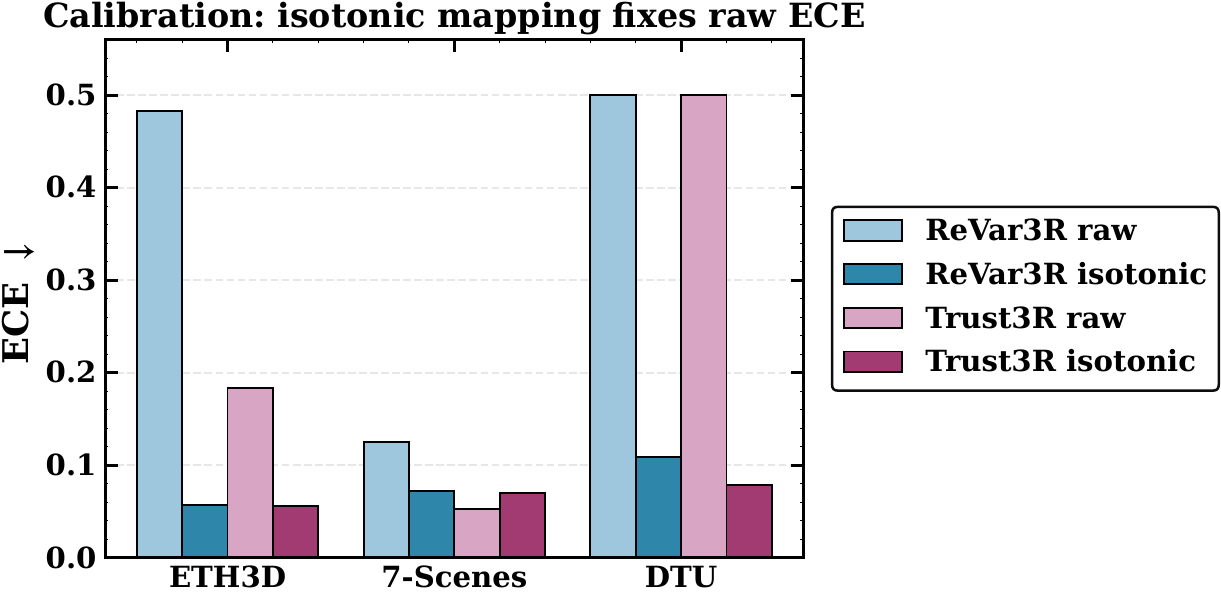}
\caption{\textbf{Calibration improves within domain.} Raw versus isotonic-calibrated ECE (\down{}) from a
within-domain calibration split. Raw perturbation scores are poorly scaled, with DTU and ETH3D saturated; a small
held-out isotonic map corrects ECE for \methodname{} and the trained head. On 7-Scenes, Trust3R's native standard
deviation is already calibrated.}
\label{fig:calibration}
\end{figure}

\begin{figure}[htb]
\centering
\includegraphics[width=0.9\linewidth]{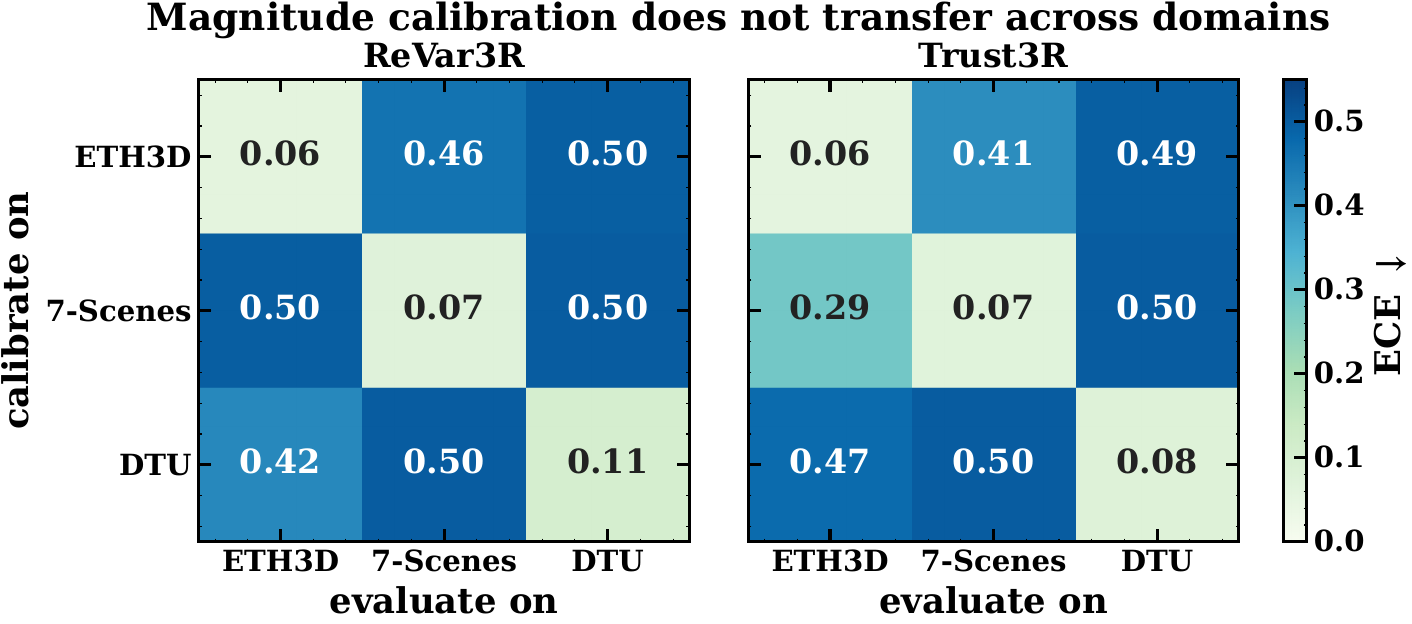}
\caption{\textbf{Calibration does not transfer across domains.} ECE (\down{}) for an isotonic map fit on the row
domain and evaluated on the column domain, for \methodname{} (left) and Trust3R (right). Both methods calibrate well
on the diagonal and poorly off it; neither transfers magnitude calibration across domains.}
\label{fig:crossdomain}
\end{figure}

\paragraph{Cross-domain calibration.} Magnitude calibration improves within domain but does not transfer. Isotonic
calibration reduces ECE on its fit domain (Figure~\ref{fig:calibration}; e.g.\ ETH3D $0.48\to0.06$), yet fitting
on one dataset and evaluating on another raises average ECE to $\approx0.46$ (Figure~\ref{fig:crossdomain}). The
transfer gap is $+0.40$ for \methodname{} and $+0.38$ for the trained head (Figure~\ref{fig:crossdomain}). Ranking
transfers; magnitude calibration does not.

\paragraph{Novel-view synthesis (NVS).} Uncertainty-guided 3D Gaussian-Splatting~\citep{kerbl2023} initialization
does not improve held-out peak signal-to-noise ratio (PSNR) under a fixed optimization budget. The five
initialization signals span $0.96$~dB, and even oracle-error initialization ($19.03$) does not exceed random
initialization ($19.52$), indicating that per-scene optimization overwrites the initialization. The estimator helps
geometric point selection, not view synthesis.

\subsection{Sample-Count Sweep}
\label{app:nsweep}
\begin{figure*}[t]
\centering
\includegraphics[width=\textwidth]{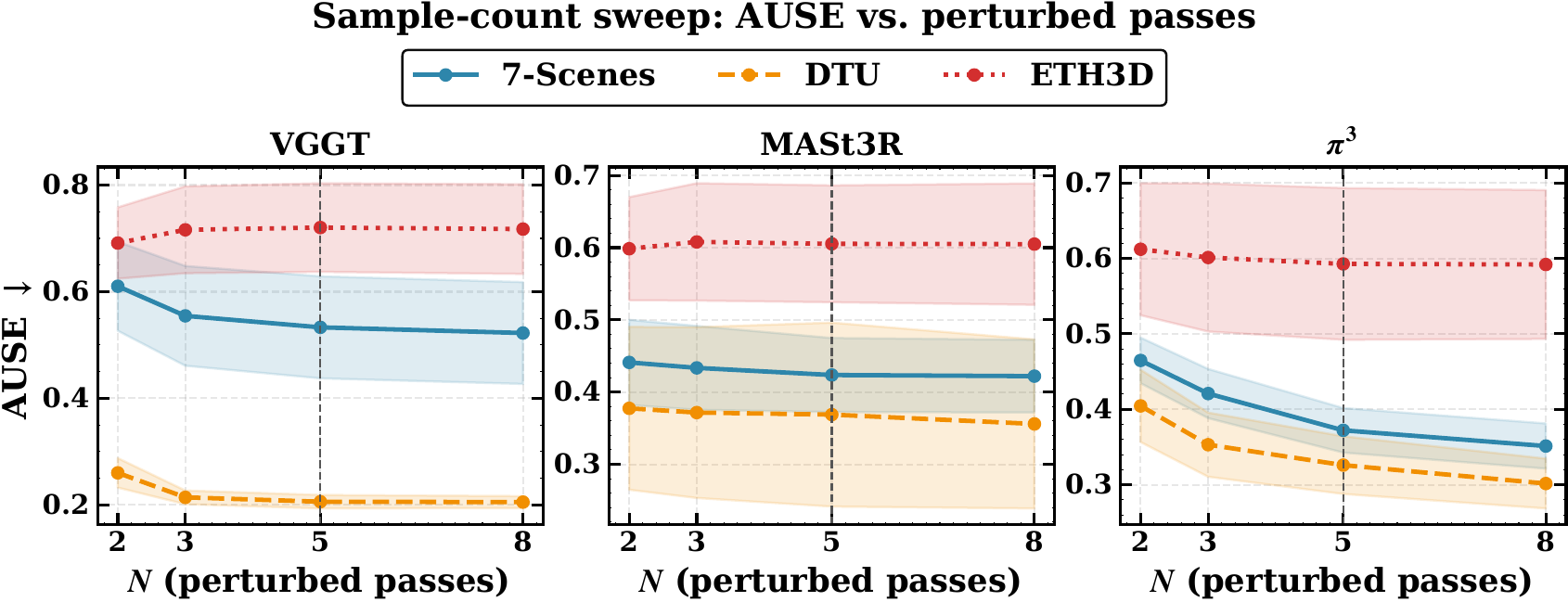}
\caption{\textbf{Sample-count sweep.} AUSE (\down{}) versus perturbed passes $N$ for each backbone and dataset;
shading gives standard deviation across view-sets, and the dashed line marks the $N{=}5$ default. In-distribution
AUSE flattens by $N{=}5$, while shifted ETH3D is nearly flat throughout.}
\label{fig:nsweep}
\end{figure*}

The default sample count lies near the in-distribution diminishing-returns knee, while shifted AUSE is largely
insensitive to sample count. Figure~\ref{fig:nsweep} varies the number of perturbed passes
$N$. In distribution, AUSE improves monotonically but flattens: $N{=}5$ captures a median $85\%$ of the improvement
available at $N{=}8$ with fewer passes, motivating the marked default. The spread across the six in-distribution
conditions is wide, $40$--$99\%$; MASt3R on DTU sits at the low end with a curve still descending at $N{=}8$, so the
default trades a little headroom there for a uniform protocol. On shifted ETH3D, AUSE remains largely
insensitive to $N$.

\subsection{Distribution-Shift Breakdown}
\label{app:ood}
Shift is not one condition, and \methodname{} does not win it uniformly. Table~\ref{tab:ood} details ETH3D on a
frozen MASt3R backbone at $224$~px over $24$ view-sets spanning nine scenes: there the trained head attains the
lowest AUSE and the best calibration, built-in confidence ranks best, and \methodname{} leads on neither. On
far-shift KITTI the ordering reverses and \methodname{} leads on both AUSE and ranking
(Section~\ref{sec:ood}). An earlier ten-view-set ETH3D sample, drawn from four scenes rather than nine, put
\methodname{} ahead on ranking and built-in confidence below random; the larger sample does not reproduce that
ordering and supersedes it. The reversal is reported rather than the smaller, more favourable sample: the honest
reading is that gauge alignment helps where the output frame moves between passes and the baseline is weak, which
is true of KITTI and not of wide-baseline ETH3D.

\begin{table*}[t]
\caption{\textbf{Distribution-shift breakdown} on ETH3D with frozen MASt3R at $224$~px, over $24$ view-sets
spanning nine scenes. Bold marks the best non-oracle value per column.}
\label{tab:ood}
\begin{center}\begin{tabular}{lccc l}
\toprule
\textbf{Method} & \textbf{AUSE~\down{}} & \textbf{Spearman~\up{}} & \textbf{ECE~\down{}} & \\
\midrule
oracle (bound) & 0.000 & 1.00 & $-$ & \\
built-in conf. & 0.585 & \textbf{0.31} & $-$ & best ranking \\
random removal & 0.743 & 0.00 & $-$ & reference \\
input-noise ensemble & 0.766 & $-0.31$ & $-$ & worse than random \\
Trust3R (trained) & \textbf{0.475} & 0.25 & \textbf{0.10} & best AUSE and calibration \\
\methodname{} & 0.667 & 0.13 & 0.38 & neither \\
\bottomrule
\end{tabular}
\end{center}
\end{table*}

\subsection{Cost and Efficiency}
\label{app:cost}
\methodname{} trades roughly ten-fold more inference passes for zero training and zero added parameters, while using
peak memory comparable to, and slightly below, the trained head. This operating point matches training-free
uncertainty estimation generally~\citep{mi2022,wang2019tta,poggi2020,ilg2018}, which replaces model adaptation with
additional inference. Table~\ref{tab:cost} compares the alternatives. It does not price a separate one-time cost per
backbone: because \methodname{} reads only outputs, it transfers to another backbone or resolution without a new
training stage. A trained head generally requires checkpoint-specific adaptation on labelled 3D data, which is why the
head-to-head in Section~\ref{sec:trust3r} is available only on MASt3R.

\begin{table*}[t]
\caption{\textbf{Cost and efficiency on the shared backbone.} \methodname{} adds no parameters or training and uses
comparable peak memory, but requires multiple passes and roughly an order of magnitude more latency than the
single-pass trained head. Latency is wall-clock time on one $16$\,GB GPU.}
\label{tab:cost}
\begin{center}\begin{tabular}{l>{\centering\arraybackslash}p{1.7cm}cccc}
\toprule
\textbf{Method} & \textbf{Training} & \textbf{Params} & \textbf{Passes} & \textbf{Peak mem.} & \textbf{Latency} \\
\midrule
Built-in conf. & none & $0$ & $0$ (free) & ${\le}2.91$\,GB & ${\sim}0.92$\,s/pair \\
Input-noise ensemble & none & $0$ & ${\sim}5$ & ${\sim}$base & subset of $9$\,s/set \\
\methodname{} & none & $0$ & ${\sim}10$--$11$ & $2.91$\,GB & $9.09$\,s/set (${\sim}9.9\times$) \\
Trust3R & $10$ ep.\ / $150$k pairs & $40.4$M & $1$ & $3.37$\,GB & $0.919$\,s/pair \\
\bottomrule
\end{tabular}
\end{center}
\end{table*}

\subsection{Additional Qualitative Results}
\label{app:qual}
\begin{figure*}[t]
\centering
\includegraphics[width=\textwidth]{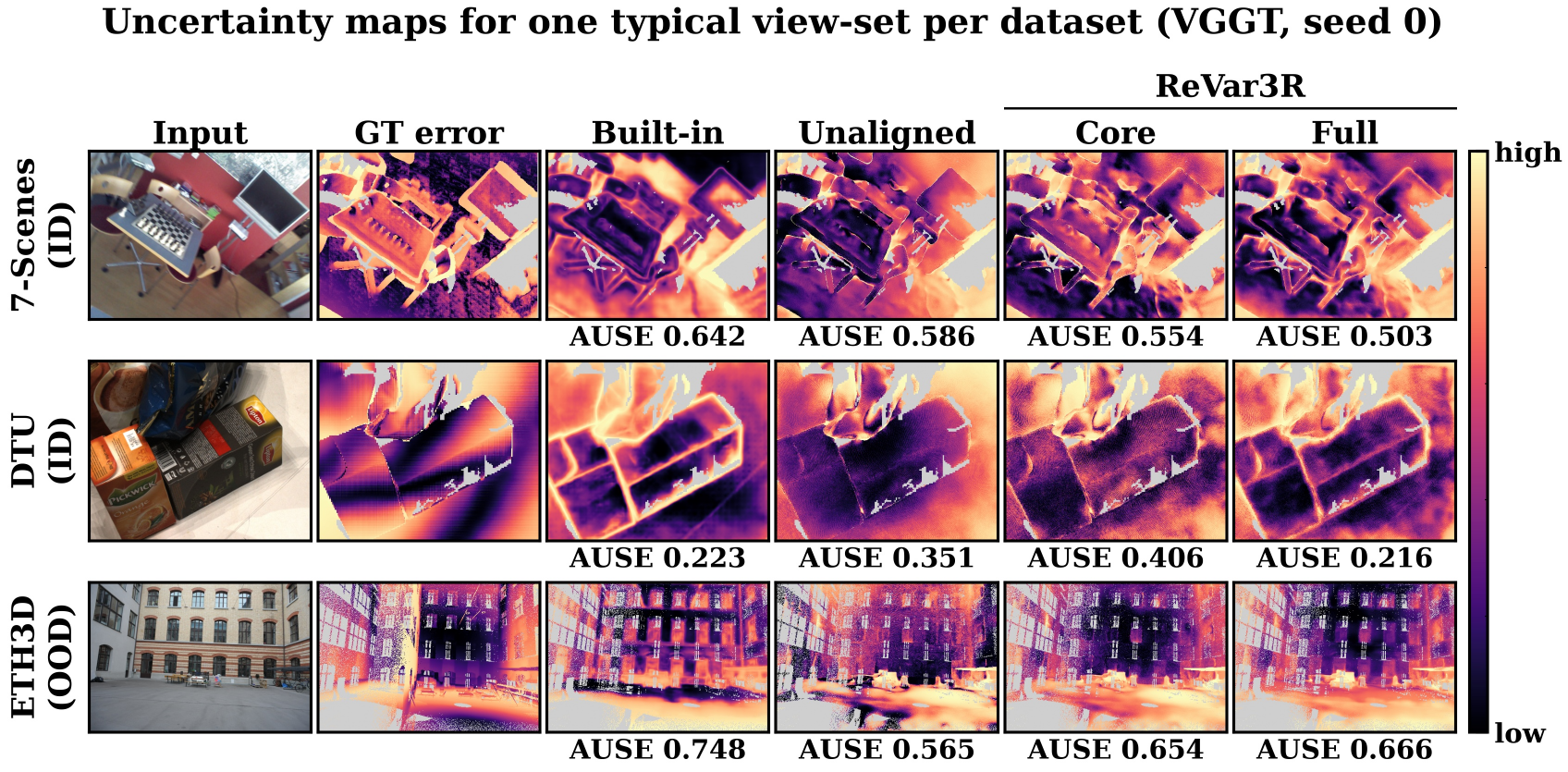}
\caption{\textbf{Qualitative uncertainty maps} from VGGT (seed 0), one typical view-set per dataset (median
within-set AUSE gap between built-in confidence and the full estimator). Columns: input, ground-truth error,
built-in confidence as uncertainty, unaligned photometric variance, and \methodname{}: the core (the same runs,
gauge-aligned) and the full estimator (held-out rank fusion with view-order variance and built-in confidence). Each
map shows every pixel's percentile rank within the view among pixels with ground truth, from low (dark) to high
(light) on the scale at the right; grey pixels have no ground truth. The number under each map is that view's own
AUSE.}
\label{fig:qual_full}
\end{figure*}

Figure~\ref{fig:qual_full} shows one typical view-set per dataset on VGGT (seed 0): the one with the median
within-set AUSE gap between built-in confidence and the full \methodname{} estimator. Each map shows percentile ranks
within the view, and the number under it is that view's own AUSE. Over each dataset's ten view-sets, the full
estimator ranks error better than built-in confidence in $9$ on 7-Scenes, $10$ on DTU and $5$ on ETH3D, and its
pooled AUSE is lower on all three ($0.491$ against $0.712$, $0.171$ against $0.248$, $0.635$ against $0.740$).
Alignment helps in distribution: the core beats unaligned variance pooled on 7-Scenes ($0.536$ against $0.655$)
and DTU ($0.236$ against $0.313$). On wide-baseline ETH3D it does not (Section~\ref{sec:alignment}), and VGGT's
unaligned variance ranks error best there (pooled $0.520$).

\section{Reproducibility Details}

\subsection{Scope of the Analysis and Design Choices}
\label{app:scope}
\paragraph{Estimated versus idealized gauge.} The proposition does not bound the residual error created by estimating
the alignment. Proposition~\ref{prop:decomp} characterizes exact removal of each run's gauge, whereas in practice
every $\hat{T}_n$ is estimated and the recovered variance retains a residual alignment-error term. Robust alignment
keeps this term small empirically: down-weighting high-residual correspondences prevents the unstable, high-error
points targeted for detection from dominating the fit and being partially reabsorbed into the gauge estimate.
Accordingly, robust Umeyama is at least as good as plain similarity fitting on every backbone
(Figure~\ref{fig:decomp}), and the recovered ranking is insensitive to the trimming fraction, Cauchy scale, and even
near-planar geometry (Appendix~\ref{app:robustalign}). A formal bound relating alignment error to outlier fraction and
scene conditioning is left to future work.

\paragraph{Reference and per-view alignment.} Reference choice matters little, but alignment scope matters. Each
perturbed run aligns to fixed reference run $0$, and each view aligns independently; the estimator is therefore best
interpreted as \emph{per-view procedural reliability} under the perturbation family, not as one scene-level
uncertainty. Table~\ref{tab:alignabl} evaluates a consensus (medoid) reference and a shared scene-level transform.
As expected under robust alignment, reference choice has little effect, whereas per-view versus scene-level scope
does. Section~\ref{sec:alignment} decomposes this difference into global-gauge and residual per-view components.

\begin{table}[tb]
\caption{\textbf{Full estimator over five seeds.} Per-condition AUSE~\down{} of the fused
estimator, mean $\pm$ standard deviation over five seeds, against built-in confidence; ``wins'' counts the
seeds in which \methodname{} is lower.}
\label{tab:seedgrid}
\begin{center}
\begin{tabular}{llccc}
\toprule
\textbf{Model} & \textbf{Dataset} & \textbf{built-in} & \textbf{\methodname{}} & \textbf{wins} \\
\midrule
\tblseedgrid
\bottomrule
\end{tabular}
\end{center}
\end{table}

\begin{table}[tb]
\caption{\textbf{Alignment robustness in simulation.} Aligned-estimator Spearman~\up{}\,/\,AUSE~\down{} against
known $\sigma_{\mathrm{true}}$ remains stable across trimming-fraction, Cauchy-scale, and near-planar sweeps. The default
drops $10\%$ and sets Cauchy scale to the median residual.}
\label{tab:robustalign}
\begin{center}\scriptsize\setlength{\tabcolsep}{3pt}\begin{tabular}{ll @{\hskip 1.8em} ll @{\hskip 1.8em} ll}
\toprule
\textbf{trim drop} & \textbf{$\rho$\,/\,AUSE} & \textbf{Cauchy $\times$} & \textbf{$\rho$\,/\,AUSE} & \textbf{depth ext.} & \textbf{$\rho$\,/\,AUSE} \\
\midrule
$0\%$ & $0.98/0.011$ & $0.25$ & $0.99/0.008$ & $2.0$ & $0.99/0.008$ \\
$5\%$ & $0.99/0.008$ & $0.5$ & $0.99/0.007$ & $1.0$ & $0.99/0.008$ \\
$10\%$ & $0.99/0.008$ & $1.0$ & $0.99/0.008$ & $0.3$ & $0.99/0.008$ \\
$20\%$ & $0.99/0.007$ & $2.0$ & $0.99/0.008$ & $0.1$ & $0.99/0.008$ \\
$30\%$ & $0.99/0.007$ & $4.0$ & $0.99/0.008$ & $0.03$ & $0.99/0.008$ \\
\bottomrule
\end{tabular}
\end{center}
\end{table}

\subsection{Robustness of the Robust Alignment}
\label{app:robustalign}

\paragraph{The internal fit does not use ground-truth validity.} The label-free alignment uses only
prediction-valid, finite correspondences; \methodname{} never sees ground truth (Section~\ref{sec:method}). As a
control, the internal fit is restricted to evaluation-valid pixels for the VGGT photometric core over ten view-sets per
dataset. Pooled AUSE changes by $-0.010$ on 7-Scenes, $-0.006$ on DTU, and $-0.088$ on sparse-ground-truth KITTI
($0.417\to0.330$). Every change is an \emph{improvement}: ground-truth validity would only help the fit. Because
the estimator does not use it, the reported gains cannot arise from ground-truth validity information, and the
label-free choice is, if anything, conservative.

The recovered ranking is insensitive to both robust-fit hyperparameters and near-degenerate geometry. The controlled synthetic harness of Appendix~\ref{app:derivation} tests trimmed
fraction, Cauchy scale, and geometry, using
gauge magnitude $0.01$ and scene distance $5$; at its defaults, the parameterized fit exactly reproduces the released
\texttt{robust\_umeyama}. Table~\ref{tab:robustalign} shows that all three sweeps preserve the ranking of known
$\sigma_{\mathrm{true}}$. Changing the trimmed fraction from $0$ to $30\%$ keeps Spearman in $[0.98,0.99]$ and AUSE
in $[0.007,0.011]$. Scaling the Cauchy width from $0.25$--$4\times$ its median-residual default changes neither past
the third decimal. 
Finally, shrinking depth extent to $1.5\%$
of lateral extent creates a near-planar scene with an ill-conditioned depth-direction similarity fit, yet ranking
remains intact ($\rho{\approx}0.99$). Lateral point spread keeps the fit well determined, so one degenerate direction
does not propagate to the estimate.

\FloatBarrier
\subsection{Seed Stability and Determinism}
\label{app:seeds}
The full $3\times6$ grid is repeated over five seeds. The fused estimator lowers AUSE below built-in confidence in
$15$ of $18$ conditions in \emph{every} seed, and no condition changes sign: each is won in five seeds out of five
or lost in five out of five, with the three losses always MASt3R and \pithree{} on BlendedMVS and \pithree{} on
ETH3D. Per-condition AUSE varies little across seeds, with a standard deviation between $0.002$ and $0.014$ (mean
$0.006$), roughly $19\times$ smaller than the median margin of $0.115$ over built-in confidence. This repetition
holds every condition at ten view-sets, whereas the main tables use eight for TUM and KITTI; the $15/18$ outcome is
identical under both protocols. The intermediate rungs are less stable than the headline: at the uniform ten-set
protocol the label-free core wins between $10$ and $12$ of $18$ depending on seed and the held-out learned fusion
between $13$ and $14$, so the stage decomposition is best read as an ordering rather than as exact counts.

A separate five-seed sweep varies alignment rather than fusion at eight view-sets per condition. There the core
lowers AUSE below built-in confidence in $11$, $11$, $10$, $10$ and $11$ of the $18$ conditions, with MASt3R on
KITTI the only condition to change sign; the core lies between $0.313$ and $0.366$ against a built-in value of
$0.346$, a margin never exceeding $0.034$ either way, so that condition is a near-tie rather than a reversal. The stochastic estimators, \methodname{} and the input-noise ensemble, are repeated over
five random seeds. By contrast, the trained Trust3R head and each backbone's built-in confidence are single
deterministic forward passes with no sampling step, so they carry no seed variation by construction.

Estimator variation is small. In the five-seed runs behind Table~\ref{tab:trust3r}, the per-dataset standard
deviation of pooled \methodname{} AUSE over seeds is $0.009$ on 7-Scenes, $0.002$ on TUM, $0.005$ on ETH3D, $0.016$ on
DTU, and $0.007$ on KITTI. The single-run headline values are therefore representative rather than fortuitous.

\end{document}